\documentclass[a4paper,twoside]{article}
\usepackage[utf8]{inputenc}
\usepackage{graphicx}
\usepackage{etex}
\usepackage[T1]{fontenc}
\usepackage[english]{babel}
\usepackage{graphicx}
\usepackage[noindentafter,pagestyles]{titlesec}
\usepackage{textcomp}
\usepackage{dcolumn}
\usepackage{subcaption}
\usepackage{amsmath}
\usepackage{algorithm}
\usepackage{algorithmic}
\usepackage{multirow}
\usepackage{enumitem}
\usepackage{placeins}  
\usepackage{float} 
\usepackage{txfonts}
\usepackage{fancyhdr}
\usepackage{threeparttable}
\usepackage{xspace}
\usepackage[superscript]{cite}
\usepackage{pictex}
\usepackage[normalem]{ulem}
\usepackage{xcolor}
\usepackage[
    colorlinks=true,
    linkcolor=blue,        
    citecolor=teal,        
    urlcolor=magenta       
]{hyperref}
\usepackage{hyperref}
\newcommand{\bmp}[1]{\begin{minipage}{#1\columnwidth}}
\newcommand{\emp}{\end{minipage}}

\newcommand{\bea}{\begin{eqnarray}}
\newcommand{\eea}{\end{eqnarray}}
\newcommand{\be}{\begin{equation}}
\newcommand{\ee}{\end{equation}}

\titleformat{\section}{\large\bfseries}{\thesection.}{.3em}{}
\titlespacing*{\section}{\leftmargini}{*3}{*3}
\titleformat{\subsection}{\bfseries}{\thesubsection}{.3em}{}
\titlespacing*{\subsection}{0pt}{*3}{*3}
\makeatletter
\def\@maketitle{%
  \newpage
  \null
  \vskip 2em%
  \begin{center}%
  \let \footnote \thanks
    {\fontsize{18}{22}\fontseries{b}\selectfont \@title \par}%
    \vskip 1.5em%
    {\normalsize
      \lineskip .5em%
      \begin{tabular}[t]{c}%
\@author
      \end{tabular}\par}%
    \vskip 1em%
    {\large \@date}%
  \end{center}%
  \par
  \vskip 1.5em}
\renewenvironment{abstract}{%
\if@twocolumn
\section*{\abstractname}%
\else
\quotation
\noindent{\bfseries\large \abstractname\vspace*{.3ex}\par}
\fi}
{\if@twocolumn\else\endquotation\fi}
\makeatother

\fancypagestyle{plain}{%
\fancyhf{} 
\fancyhead[L]{\small \MakeUppercase{\hspace*{15mm} 11$^{\mbox{\tiny th}}$ European Conference for Aeronautics and AeroSpace Sciences (EUCASS)}}
\fancyfoot[L]{\small Copyright \copyright\ 2025 by NEUHALFEN, GRZYMISCH, S.-GESTIDO. Published by the EUCASS association with permission.}%
} 
\begin{document}
%
%
%
%
%

\title{Enabling Robust, Real-Time Verification of Vision-Based Navigation through View Synthesis}

\author{%
  \begin{tabular}{ccc}
    Marius Neuhalfen$^{1,2,3\dagger}$\footnote{Contact: \textit{neuhalfen [dot] marius [at] gmail [dot] com}. This work was carried out while the corresponding author was a trainee at ESA.} &
    Jonathan Grzymisch$^{1}$ &
    Manuel Sánchez-Gestido$^{1}$ \\
  \end{tabular}\\[1ex]
  \begin{tabular}{ccc}
    \multicolumn{3}{c}{$^1$\textit{European Space Agency, ESTEC, The Netherlands}}\\
    \multicolumn{2}{c}{$^2$\textit{RWTH Aachen University, Aachen, Germany}} &
    \multicolumn{1}{c}{$^3$\textit{École Centrale de Lille, Lille, France}}\\[1ex]
    \multicolumn{3}{c}{$^\dagger$\textit{Corresponding author}}
  \end{tabular}%
}

\date{\small{\text{June 15, 2025}}}

\maketitle

\vspace*{-1cm}

\begin{abstract}
  \noindent
This work introduces VISY-REVE: a novel pipeline to validate image processing algorithms for Vision-Based Navigation. Traditional validation methods such as synthetic rendering or robotic testbed acquisition suffer from difficult setup and slow runtime. Instead, we propose augmenting image datasets in real-time with synthesized views at novel poses. This approach creates continuous trajectories from sparse, pre-existing datasets in open or closed-loop. In addition, we introduce a new distance metric between camera poses, the Boresight Deviation Distance, which is better suited for view synthesis than existing metrics. Using it, a method for increasing the density of image datasets is developed. Project page\footnote{Project Page Link: \href{https://marius-ne.github.io/visyreve.github.io/}{https://marius-ne.github.io/visyreve.github.io/}}
\end{abstract}

\section{Introduction}
Autonomous Guidance, Navigation and Control (GNC) technologies are becoming increasingly relevant for modern space applications. Such systems must be able to independently make decisions, plan maneuvers, and safely execute them; all without the intervention of operators. Two prominent mission profiles that need these capabilities are in-orbit close-proximity operations with human-made targets~\cite{Schnitzer2017} (e.g., satellite servicing or debris removal) and entry, descent, and landing on natural bodies~\cite{Delaune2021} (e.g., the Moon or Mars).  
These autonomous GNC capabilities often rely on \emph{vision-based navigation} (VBN)~\cite{DAmico2014}, where images from passive or active optical sensors provide relative state information between a target and the host spacecraft. Due to the high reliability requirements of space missions, the entire GNC stack, including the VBN, must be carefully validated on the ground before launch. This validation includes testing the image processing algorithms that are part of the VBN, ideally with hardware in the loop and in real-time~\cite{fernandezphobos}.

Unfortunately, representative on-orbit imagery necessary validation (examples include LIRIS~\cite{Masson2017} and PRISMA~\cite{DAmico2014,bodin2012}) is scarce, often proprietary, and rarely accompanied by accurate ground truth. \textit{In-situ} testing before a mission is impossible, and transferring results from previous datasets is unreliable. Therefore, VBN validation relies on surrogate imagery that approximates the in-space environment as closely as possible, ideally minimizing the domain gap~\cite{cassinis2022domaingap}.

\section{Related Work}
\textbf{Surrogate Validation Datasets for VBN} \\
Currently, two surrogate image acquisition methods to generate validation datasets dominate. 

The first is \textit{synthetic rendering}. Here, software packages such as PANGU~\cite{martin2019pangu} or SurRender~\cite{Lebreton2021} aim to recreate the real environment of space and to model radiometric properties from first principles. A particular advantage of this method is that it can simulate a variety of optical effects, which is useful for Monte Carlo campaigns and robustness testing of image processing algorithms. However, creating accurate reflection models, high-fidelity 3D geometries and material textures requires a significant upfront effort. Additionally, rendering of accurate synthetic imagery remains computationally intensive~\cite{Lebreton2021} and therefore difficult to integrate into real-time testing, particularly on flight hardware.

The second uses \textit{robotic testbeds (R.T.)} (also called \textit{laboratories}). These are special facilities where physical spacecraft mock-ups are manipulated with robotic actuators and photographed under calibrated lighting. The advantage of this approach is that the images will include the effects of a real optical sensor, including lens distortions, noise and other artifacts. However, the space of possible poses (positions and attitudes) is inherently limited, mock-ups must be scaled down to fit inside the facility, which leads to deviations in lighting, materials must be recreated or approximated, and each campaign requires careful re-calibration and integration with facility actuators and processors~\cite{Cassinis2020,CassinisPHD2022}. 

Combining synthetic and R.T. methods is advantageous as it has been shown that training models with mixed imagery from both domains increases the domain adaptation of the models from the surrogate to in-situ images~\cite{Kisantal2020,Schnitzer2017}.

Datasets can further be divided into \emph{trajectory-based} datasets that use physically motivated camera paths, and \emph{random-pose} datasets that randomly sample camera poses. Trajectory-based datasets are preferable for open-loop tests, but generally cover the whole image space less well than random-pose datasets without additional effort.

Our method of \textit{view synthesis} is complementary to these surrogate image acquisition methods; it can synthesize images in real-time from an existing dataset, be it R.T. or synthetic. It can create new images similar in quality to the input images without having to perform heavy rendering or complex interfacing with a facility. This makes it possible to enhance previously acquired datasets, yielding more imagery, and enabling them to be expanded in real-time reacting to navigation results from a GNC stack, thus enabling closed-loop functional testing of image processing.

\textbf{View Synthesis} \\
\label{relatedwork:VS}
The term View Synthesis (VS) refers to the process of generating novel imagery from an existing sparse set of images acquired by cameras located at different viewpoints (poses)~\cite{CANCLINI201940}. This is achieved by deriving underlying information about the scene from the input images and using this information to transform them to obtain novel views of the scene from different camera poses. We refer to the combination of an image and its associated camera pose as a \textit{view}.

VS has been used in the space domain by~\cite{Perez2024} for self-supervised test-time adaptation of convolutional neural networks (CNN) for keypoint-based pose estimation. The authors showed that VS is beneficial to improve the CNN's ability to bridge the domain gap, that is, the difference between the aforementioned surrogate image types and \textit{in-situ} imagery. However, the synthesized views needed to be temporally close to their source, which is not the case in our work.

In the context of validation, VS has been used in other domains such as autonomous driving, for example by VISTA~\cite{Amini2020,DeOliveira2021}, to validate control of self-driving cars using sparsely imaged trajectories. The authors showed in a similar manner as before that training models with off-line collected sparse datasets together with in-between views synthesized by VS in real-time improves the transfer capabilities of these models to real-world driving scenarios.

\textbf{Pose Distance Metrics} \\
VS techniques often involve computing distances between camera poses. This is done, for example, when choosing an optimal source image from multiple near-neighbor candidates. The more distant (loosely defined) the novel camera pose is from the one of the chosen source view, the more difficult is the VS and the less accurate will be the result. The difficulty arises from the fact that, in general, VS struggles to synthesize novel information~\cite{mildenhall2020nerfrepresentingscenesneural}. The novel information in this case stems from novel parts of the scene becoming visible when the pose of the camera changes. 

\begin{figure}[t]
    \centering
    \makebox[\textwidth][c]{%
        \includegraphics[width=1.1\linewidth]{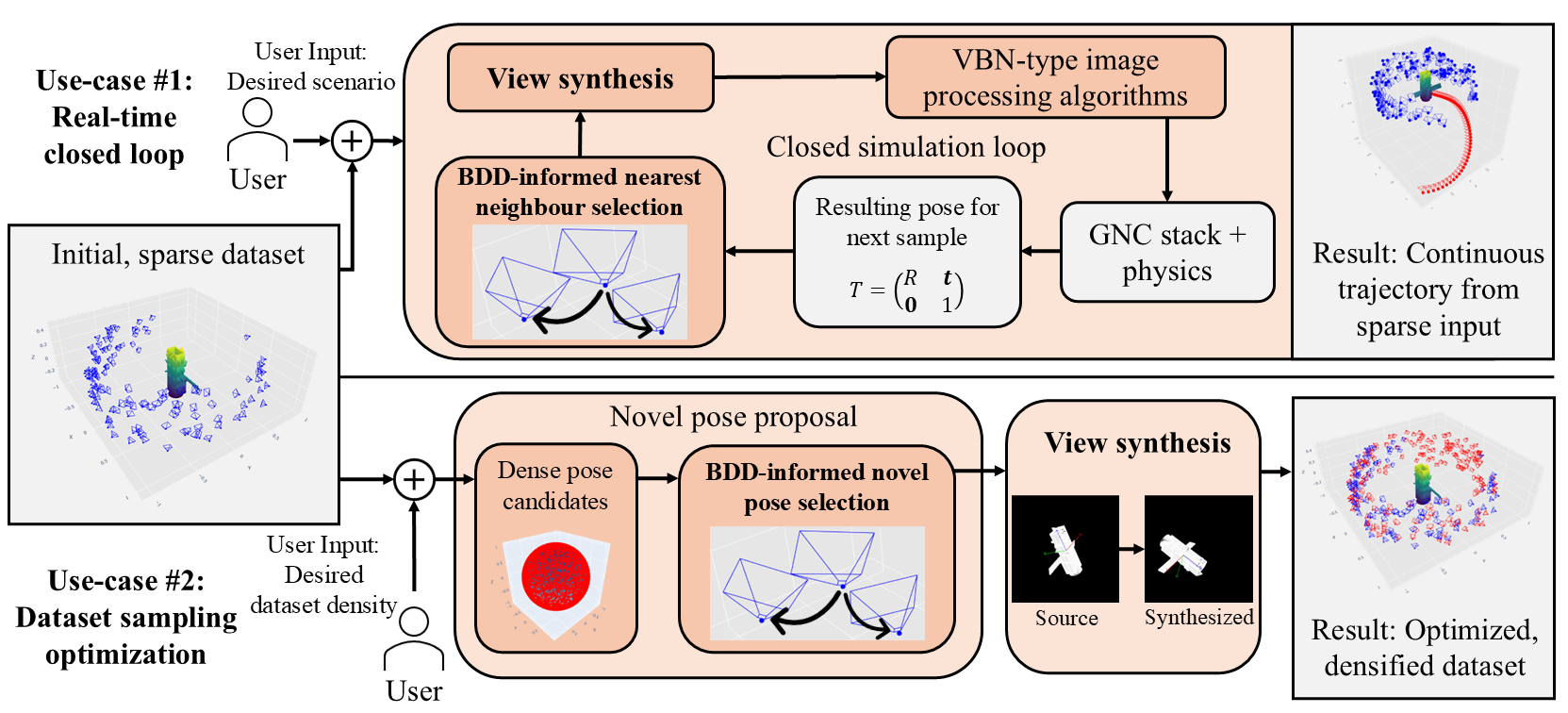}
    }
    \caption{\textbf{Key contributions of this work}. We introduce a versatile view synthesis pipeline that fits a variety of use-cases. Two example ones are illustrated here: 1. Running closed-loop, real-time tests of VBN image processing algorithms from an existing, \textit{sparse} dataset. 2. Optimizing pose sampling of a dataset to increase image density, note that this can also be used in advance of dataset creation to efficiently plan image acquisition campaigns.}
    \label{fig:keycontribs}
\end{figure}

Thus, it becomes necessary to quantify the distance between camera poses, both to choose source views and to create performance models that indicate the limits of VS. For this purpose, multiple pairwise distance metrics exist; widely used is the Euclidean norm of the displacement between camera positions \cite{xu2021layoutguidednovelviewsynthesis,wang2024freevsgenerativeviewsynthesis,Amini2020}. 

Alternatively, one can use the magnitude of the relative rotation between two views~\cite{Liu_2022}. This better encapsulates the fact that changes in the visible parts of a scene can result even from translationally static cameras. However, it ignores the fact that new parts of a scene can also become visible by pure translation of the camera.

To resolve these shortcomings, we propose a new metric to measure the distance between camera poses that is able to better predict the difficulty of view synthesis than the two previously mentioned metrics. We call it the \textit{Boresight Deviation Distance}. Its definition will be given later on.

\subsection{Our Contributions}
This work introduces a real-time VS pipeline that generates novel views from \textit{a priori} available sparse image datasets with associated ground-truth pose labels. We call this pipeline \textbf{VISY-REVE} which stands for \textbf{VI}ew \textbf{SY}nthesis for \textbf{RE}al-time \textbf{V}alidation for space \textbf{E}xploration. The novel views can be used to run closed-loop testing campaigns with imagery that responds in real-time to changes in the pose of a simulated camera. Additionally, it can be used off-line to enhance the image sampling of a dataset. These applications are illustrated in~\ref{fig:keycontribs}. To the best of our knowledge, our work is the first to introduce this technique to the problem of real-time validation in the space domain. 

Our approach closes the gap between photorealistic but slow rendering for software-in-the-loop and real-time but difficult-to-set up R.T. acquisition for hardware-in-the-loop validation. VISY-REVE's two key contributions are:

\begin{enumerate}
    \item \textbf{Real-time view-synthesis pipeline.}  
          The pipeline accepts sparse input datasets (e.g. synthetic or robotic-testbed, optionally including depth or 3D model) and synthesizes geometrically accurate novel views at real-time speed.

    \item \textbf{Boresight Deviation Distance.}  
          A novel distance metric between camera poses that scales well with VS quality, that is able to agnostically compare different datasets and that can identify poorly sampled pose regions.
\end{enumerate}

\section{Conventions}
\label{sec:conventions}
Our work relies on the notions of projective geometry and related concepts from computer vision. A detailed description of these topics can be found in~\cite{Hartley2013multipleviewgeometry}. Here we will only give a short overview of the notation used. The projection of a 3D point given in target coordinates into an image is given as:

\begin{align}
    \begin{bmatrix}
        \hat{u} \\
        \hat{v} \\
        \hat{z}
    \end{bmatrix} = K \ [I \ | \ \textbf{0}] \ T_{target2cam} \ \begin{bmatrix}
        x \\
        y \\
        z \\
        1
    \end{bmatrix}
\end{align}

Where both points are given in homogeneous coordinates. The 2D image coordinates can therefore be obtained by dividing both $\hat{u}$ and $\hat{v}$ by $\hat{z}$. The projection matrix is given with the extrinsics $T_{target2cam}$ and intrinsics $K$. Extrinsics are given as a passive target to camera rigid body transformation. The boresight (camera) axis is defined to be looking down the positive $z$ direction. The intrinsics are based on a pinhole camera model with zero skew. Formally:

\begin{align}
\label{intrinsics}
    K\ =\begin{bmatrix}
        f_x & 0 & p_x \\
        0 & f_y & p_y \\
        0 & 0 & 1
    \end{bmatrix}, \
    T_{target2cam} = 
\begin{bmatrix}
  R & \mathbf{t} \\[6pt]
  \mathbf{0}^T & 1
\end{bmatrix}
\end{align}

We refer to mappings between images as image transformations. These transformations operate on each pixel in the input image and map it to a given pixel in the output image. Some transformations do this smoothly across the image; in this case, the transformations are affine or projective and can be expressed by $3\times3$ matrices in the homogeneous image coordinates. Other transformations are non-linear warps that act on each pixel independently.

\section{Datasets}
To use VS, preexisting datasets are needed. We rely exclusively on publicly available ones for the reproducibility of the achieved results. Note that this work focuses exclusively on the human-made target scenario although we acknowledge that the methods presented can also be applied to the natural target scenario.

The following datasets are chosen:
\begin{enumerate}
    \item SPEED+ dataset (synthetic only) \cite{Park2021}, \href{https://purl.stanford.edu/wv398fc4383}{Link}
    \item SWISSCUBE dataset \cite{Hu2021}, \href{https://huggingface.co/datasets/EPFL-CVLAB-SPACECRAFT/SwissCube/blob/main/readme.txt}{Link}
    \item ESA-AIRBUS dataset (MAN-DATA-L1) \cite{Lebreton2024}, \href{https://vbntrainingdatasets.esa.int/rendezvous/#man-data-l1_envisat_lab_mockup}{Link}
\end{enumerate}

\begin{figure}[t]
  \centering
  \begin{minipage}[b]{0.25\textwidth}
    \centering
    \includegraphics[width=\textwidth]{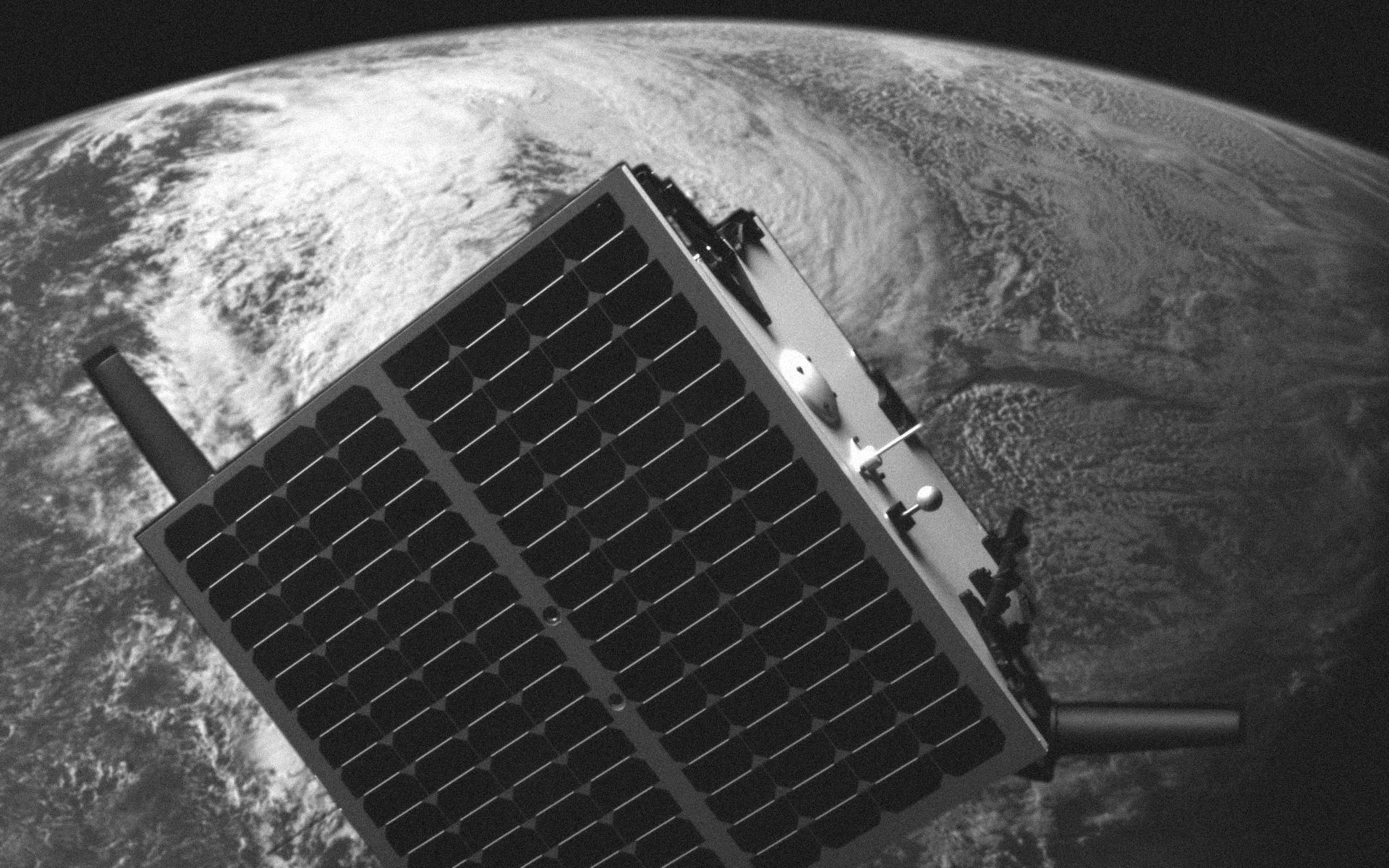}
    \caption{SPEED+}
  \end{minipage}%
  \hfill
  \begin{minipage}[b]{0.25\textwidth}
    \centering
    \includegraphics[width=\textwidth]{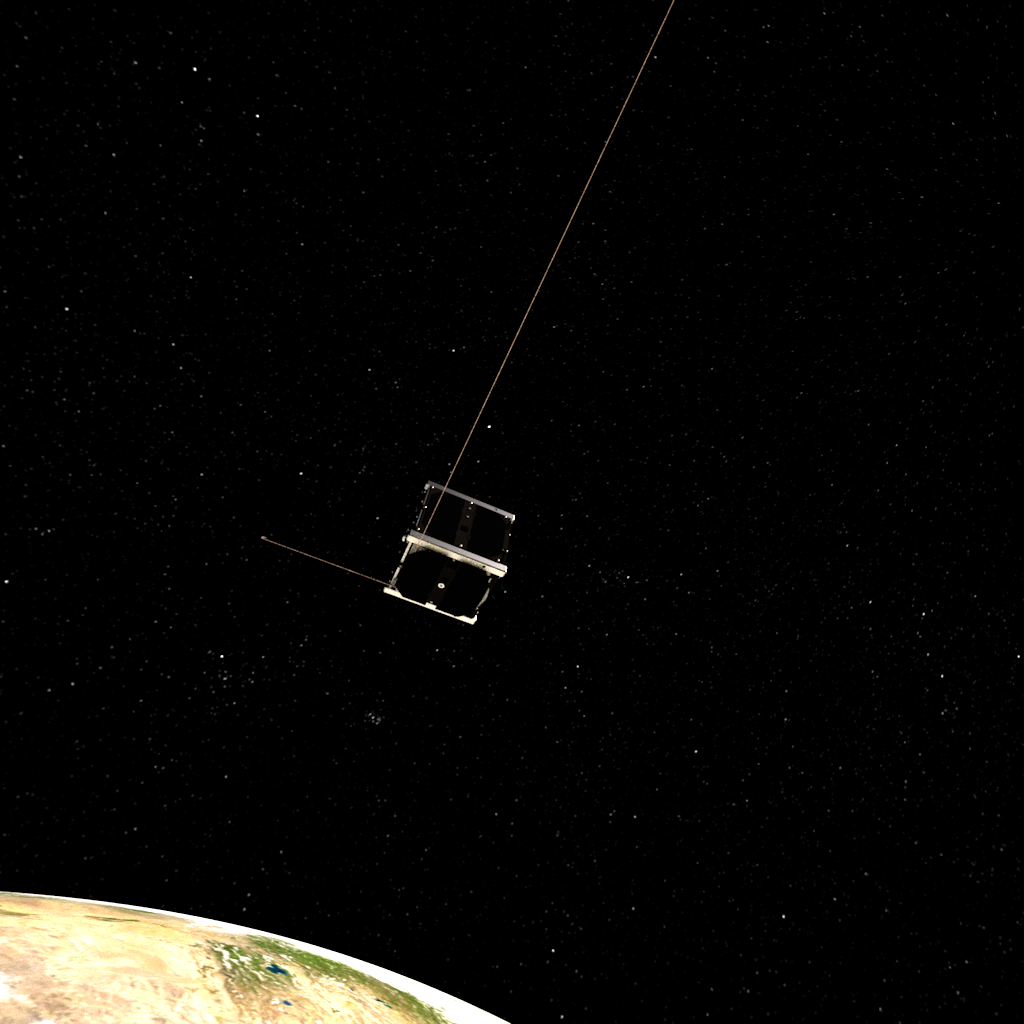}
    \caption{SWISSCUBE}
  \end{minipage}%
  \hfill
  \begin{minipage}[b]{0.25\textwidth}
    \centering
    \includegraphics[width=\textwidth]{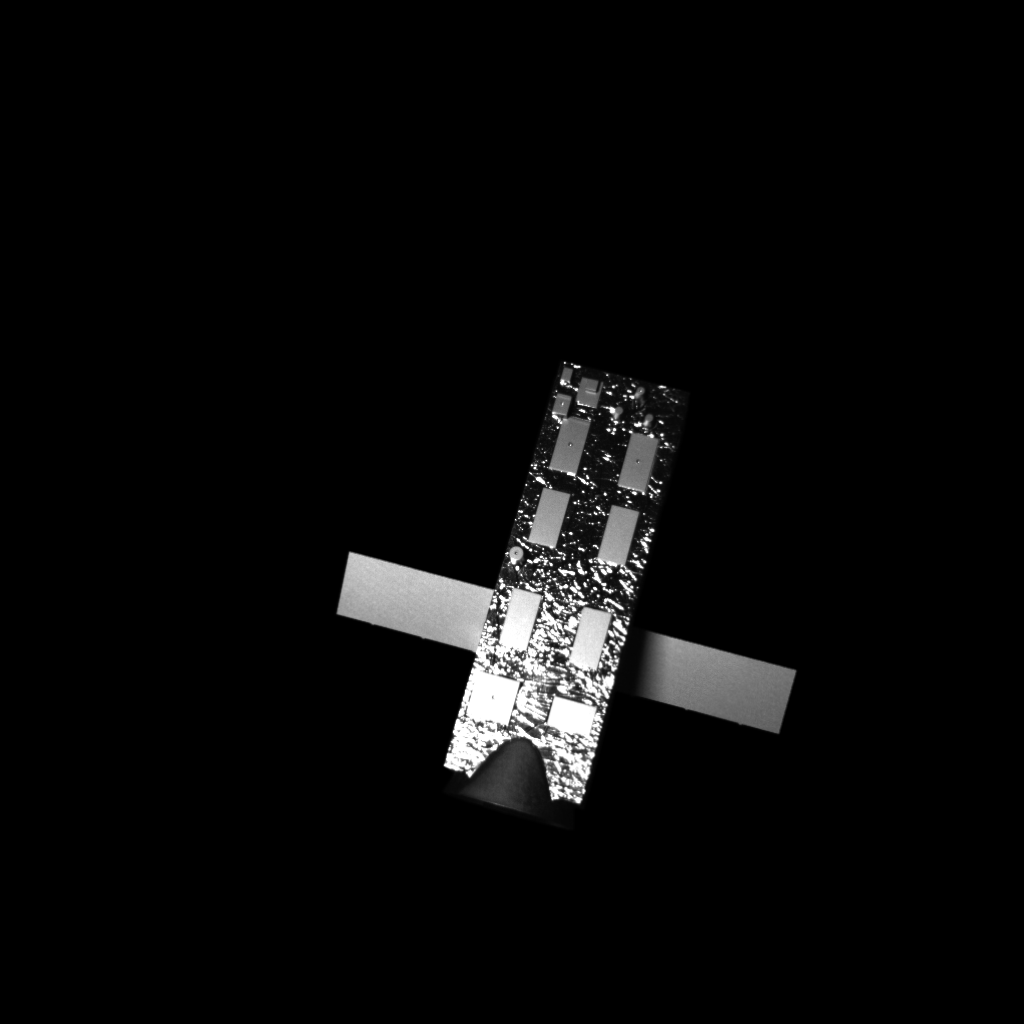}
    \caption{ESA-AIRBUS}
  \end{minipage}%
  \caption{Examples from the chosen datasets.}
  \label{fig:dataset_examples}
\end{figure}

See ~\ref{fig:dataset_examples} for example images from  these datasets. Below is a short table that summarizes their key characteristics.
    
\begin{table}[ht]
\setlength{\tabcolsep}{8pt}
  \centering
  \caption{Comparison between the chosen datasets}
  \label{tab:dataset_comparison}
  \begin{tabular}{l|ccccc}
    \hline
    \textbf{Dataset} & \textbf{Image-type} & \textbf{Poses} &\textbf{Color} & \textbf{Num. of Images} & \textbf{Target Satellite} \\
    \hline
    SPEED+ & R.T. \& Synthetic & Random & Grayscale & 9500 (R.T.), 60.000 (S) & Tango \\
    SWISSCUBE & Synthetic & Trajectories & RGB & 43.209 & Swisscube \\
    ESA-AIRBUS & R.T. & Random & Grayscale & 16.000 & Envisat \\
    \hline
  \end{tabular}
\end{table} 

The ESA-AIRBUS and SWISSCUBE datasets include the 3D models of the target mock-ups used. 

For the SPEED+ dataset, the 3D model was not publicly available at the time of writing. To still be able to apply model-based methods, we reconstruct a coarse 3D \texttt{.obj }model from the dataset's images and their associated ground-truth labels. We release it alongside this work\footnote{\href{https://github.com/marius-ne/Tango3DModel}{https://github.com/marius-ne/Tango3DModel}}. This model is equally useful for other datasets that rely on the same mock-up, namely SPEED \cite{Sharma2019} and SHIRT \cite{Park2022}.

\section{Boresight Deviation Distance}
In this chapter, we introduce a novel pose distance metric with the goal of characterizing the performance of VS. First, we motivate the need for its definition intuitively (\ref{sec:bddmotivation}) and formally (\ref{sec:bddproof}). We state its definition (\ref{sec:bdddefinition}), derive its useful properties (\ref{sec:bddadvantages}) and use it to define a metric of sampling density of datasets (\ref{sec:Density}). 

\subsection{Motivation}
\label{sec:bddmotivation}
To create a performance model for VS, it is necessary to have a pairwise distance metric between poses that is \textbf{as proportional as possible} to the error in the synthesized images. This is because the performance model is based on the $3\sigma$ upper bound of an empirically derived relationship between distance and error. Therefore, the more proportional the error is to the distance metric, the smaller the uncertainty, and therefore the more accurate the performance model.

The distance metric usually chosen for this task is the L2-norm of the distance between the camera positions in $\mathbb{R}^3$, i.e.\ $||\mathbf{t_i} - \mathbf{t_j}||$ or $||\mathbf{C}_i-\mathbf{C}_j||$ where $\mathbf{C} = -R^{-1} \ \mathbf{t}$. In this work, we refer to this metric as \textbf{C-L2}. In the following, we show that it is not an ideal choice. Its main flaw is the fact that it is $isotropic$ w.r.t. the axes of the camera reference frame, i.e. that it treats a translation in e.g. the $x$ and $z$ axes in the same way. This makes it weakly proportional to the error.

\begin{figure}[t]
    \centering
    \begin{subfigure}{0.34\textwidth}
        \includegraphics[width=\linewidth]{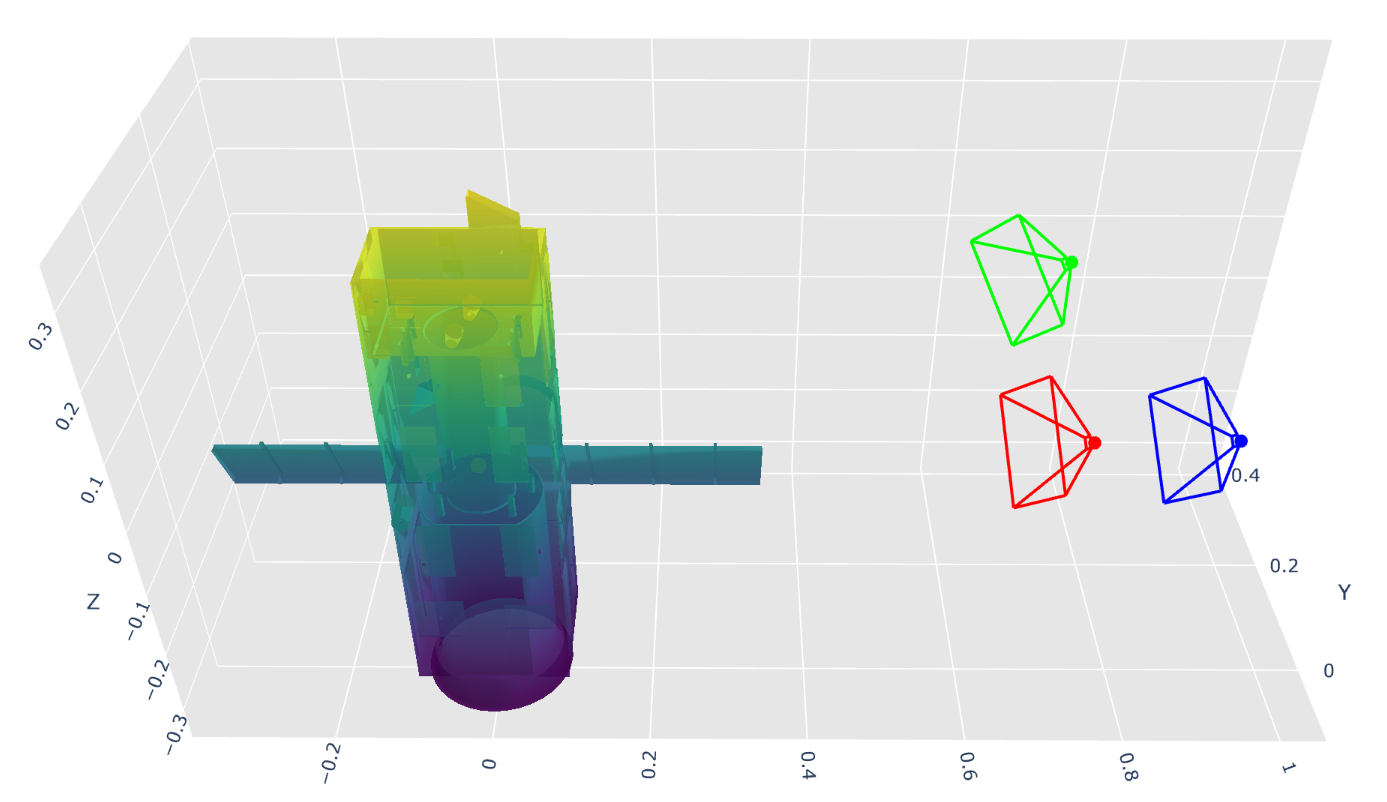}
        \caption{Camera poses from right in 3D.}
        \label{fig:bddspace}
    \end{subfigure}
    \hfill
    \begin{subfigure}{0.65\textwidth}
        \includegraphics[width=\linewidth]{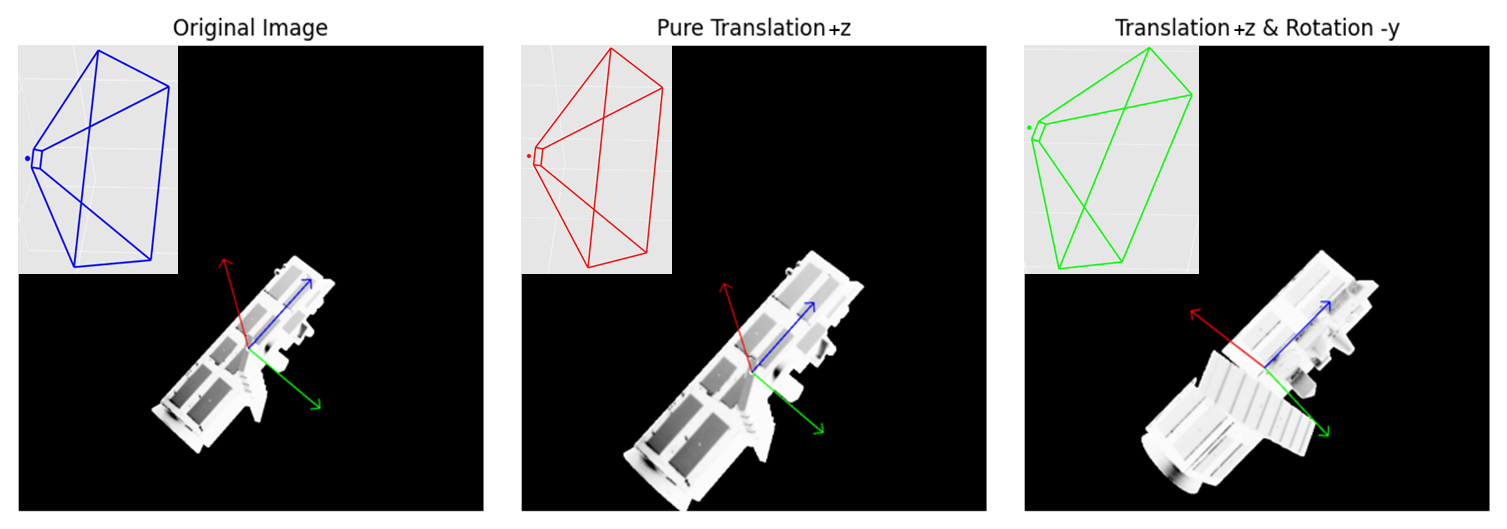}
        \caption{Images from each pose.}
        \label{fig:bddrender}
    \end{subfigure}
    \caption{Comparison of poses with similar C-L2 but different BDD values. The green view (right) is harder to synthesize from the blue view (left) due to both translation in $z$ and rotation in $y$ of the camera. Compared to the green view, the red view (center) is perceptually closer to the blue view due to only being translated in $z$, despite both the red and green views having a similar C-L2 distance to the blue view. This anisotropy is captured by the BDD.}
    \label{fig:bdd_comparison}
\end{figure}

To solve this, we introduce the \textbf{Boresight Deviation Distance} (BDD), a novel distance metric between poses, more specifically, their attitudes. Intuitively, the BDD measures how far the rotation axis of the relative rotation between two poses is from the camera’s viewing direction (boresight axis). The basis for its definition is the simple observation that the quality of VS depends on the relative pose in an $anisotropic$ manner. The relative pose is the displacement of the camera from the pose of a source view to the one of a target view. Some relative poses are easier to synthesize using VS than others. This is shown visually in~\ref{fig:bdd_comparison}. Although this is intuitive, we want to prove it formally.

Specifically, we investigate which image transformations depend on having precise depth information to the individual corresponding 3D points and which do not. If knowledge of pixel-wise depth information is necessary, then VS becomes more difficult, as this depth information is lost by projection, as shown in~\ref{sec:conventions}.

Therefore, we investigate which components of a relative pose imply an image transformation that needs precise depth information and which do not.

\subsection{Proof of the Anisotropy of View Synthesis}
\label{sec:bddproof}
\textbf{Notation:} In the following $A$ (upper-case) will refer to a matrix and $\mathbf{a}$ (lower-case, bold) to a vector. Specifically, $\mathbf{x}$ will refer to a 2D image point, $\mathbf{x_{hom}}$ to its homogeneous counterpart, $\mathbf{p_C}$ to the corresponding 3D point and $\mathbf{\hat{(\,\cdot\,)}}$ to the transformed version of a point. All 3D points and axes are given in the reference frame of the camera.

First, we will derive the expression for the image transformation associated with a pure relative rotation around the $z$ axis of the camera (boresight axis). We expect that this transformation is possible to perfectly recreate using VS as it is simply a planar rotation of the image. We will choose an arbitrary image point  $\mathbf{x} = [u, v]$ with its homogeneous counterpart defined as $\mathbf{x}_{\text{hom}} = \lambda \ \begin{bmatrix}u & v & 1\end{bmatrix}^T$. This is the ray extending from the camera center $\mathbf{C}$ through the image point $\mathbf{x}$. Any point along this ray will project to the image point. This fact is represented by the scale factor $\lambda$. 

The homogeneous image point is first back-projected into camera space using the inverse of the intrinsic matrix: 

\begin{equation}
    \mathbf{p}_c
  = \lambda \ K^{-1} \ \mathbf{x}_{\text{hom}}
  = \lambda\begin{bmatrix}
          \dfrac{u-p_x}{f_x}\\
          \dfrac{v-p_y}{f_y}\\
          1
        \end{bmatrix}
\end{equation}

\noindent
Then, a relative rotation of $\alpha$ around the boresight axis $z$ is applied:

\begin{figure}[h]
  \centering
  \begin{minipage}{0.48\linewidth}
    \begin{equation}\label{eq:hat_p}
      \mathbf{\hat{p}_C}
      = R\,\mathbf{p_C}
      =
      \begin{bmatrix}
         \cos\alpha & -\sin\alpha & 0\\
         \sin\alpha &  \cos\alpha & 0\\
         0          &  0          & 1
       \end{bmatrix}
       \;\lambda\;
       \begin{bmatrix}
        \dfrac{u-p_x}{f_x}\\[6pt]
        \dfrac{v-p_y}{f_y}\\
        1
       \end{bmatrix}
    \end{equation}
  \end{minipage}
  \begin{minipage}{0.48\linewidth}
    \begin{equation}\label{eq:hat_x}
      \mathbf{\hat{x}_{hom}}
      = K\,\mathbf{\hat{p}_C}
      = K\,R\,\lambda
      \begin{bmatrix}
        \dfrac{u-p_x}{f_x}\\[6pt]
        \dfrac{v-p_y}{f_y}\\
        1
      \end{bmatrix}
    \end{equation}
  \end{minipage}
\end{figure}

\noindent
The rotated 3D point in the camera space is then projected into the target image:

\begin{figure}[h]
  \centering
  \begin{minipage}{0.48\linewidth}
    \begin{equation}\label{eq:hat_xhom}
      \mathbf{\hat{x}_{hom}}
      = \lambda
      \begin{bmatrix}
        f_x\bigl(\cos\alpha\,\bar u - \sin\alpha\,\bar v\bigr) + p_x\\[6pt]
        f_y\bigl(\sin\alpha\,\bar u + \cos\alpha\,\bar v\bigr) + p_y\\[6pt]
        1
      \end{bmatrix}
      = \lambda \, A\,\mathbf{x_{hom}}
    \end{equation}
  \end{minipage}
  \begin{minipage}{0.48\linewidth}
    \begin{equation}\label{eq:bar_uv}
      \bar u = \frac{u - p_x}{f_x}, \quad
      \bar v = \frac{v - p_y}{f_y}
    \end{equation}
  \end{minipage}
\end{figure}

As shown, the homogeneous transformed image point can be obtained by an \textit{affine} transform $A$ of the original homogeneous image point. The transformed 2D image point is obtained by dividing the homogeneous point by its $z$ coordinate, which cancels the scale factor $\lambda$.

\begin{equation}
    \mathbf{\hat{x}} = 
    \begin{bmatrix}
        \hat u\\ 
        \hat v
    \end{bmatrix} = 
    \begin{bmatrix}
         \cos\alpha\,u
         - \sin\alpha\,\dfrac{f_x}{f_y}\,v
         + p_x(1-\cos\alpha)
         + \sin\alpha\,\dfrac{f_x}{f_y}\,p_y \\
         \sin\alpha\,\dfrac{f_y}{f_x}\,u
         + \cos\alpha\,v
         + p_y(1-\cos\alpha)
         - \sin\alpha\,\dfrac{f_y}{f_x}\,p_x
    \end{bmatrix}
    \iff \mathbf{\hat{x}} \;\neq\; f(\lambda)
\end{equation}

This shows that a relative rotation exclusively around the boresight axis can be represented with an affine matrix without the need for precise depth information. Crucially, this means that it can be perfectly recreated with VS.

Next, we proceed in a similar way for a translation in the direction of $z$ (boresight axis). For the sake of demonstration, we also include a translation in the $x$ and $y$ axes; in these axes, however, the translation must be proportional to the depth $\lambda$ of the corresponding 3D point as we will see below. In general, the image transformation associated with an arbitrary translation in $x$ and $y$ is depth-dependent. Therefore, for the proof, we model the translations in $x$ and $y$ as linearly proportional to the depth with a constant factor $\beta$, so $\Delta x = \beta_x \ \lambda$ and $\Delta y = \beta_y \ \lambda$. 

\begin{figure}[H]
  \centering
  \begin{minipage}{0.4\linewidth}
    \begin{equation}\label{eq:hat_pC}
      \hat{\mathbf{p}}_C
      = \mathbf{p}_C + \Delta\mathbf{p}
      = \begin{bmatrix}
        \dfrac{\lambda}{f_x}\,(u - p_x) + \beta_x\,\lambda \\[6pt]
        \dfrac{\lambda}{f_y}\,(v - p_y) + \beta_y\,\lambda \\[6pt]
        \lambda + \Delta z
      \end{bmatrix}
    \end{equation}
  \end{minipage}
  \begin{minipage}{0.55\linewidth}
    \begin{equation}\label{eq:hat_xhom}
      \hat{\mathbf{x}}_{\mathrm{hom}}
      = K\,\hat{\mathbf{p}}_C
      = \begin{bmatrix}
        \lambda\,(u - p_x) + f_x\,\beta_x\,\lambda + p_x\,(\lambda + \Delta z) \\[6pt]
        \lambda\,(v - p_y) + f_y\,\beta_y\,\lambda + p_y\,(\lambda + \Delta z) \\[6pt]
        \lambda + \Delta z
      \end{bmatrix}
    \end{equation}
  \end{minipage}
\end{figure}

\noindent
Again, we divide by the homogeneous component to obtain the transformed image point:

\begin{equation}
    \mathbf{\hat{x}} =
    \; \frac{1}{\,\lambda + \Delta z\,} \;
    \begin{bmatrix}
        \lambda\,(u - p_x) \;+\; f_x\,\beta_x \, \lambda \;+\; p_x (\lambda + \Delta z) \\
        \lambda\,(v - p_y) \;+\; f_y\,\beta_y \, \lambda \;+\; p_y (\lambda + \Delta z)
    \end{bmatrix}
    \;=\;
    \begin{bmatrix} 
        \tfrac{\lambda}{\lambda + \Delta z}\,(u-p_x) \;+\; \tfrac{\lambda}{\lambda + \Delta z} f_x\,\beta_x \;+\; p_x \\
        \tfrac{\lambda}{\lambda + \Delta z}\,(v-p_y) \;+\; \tfrac{\lambda}{\lambda + \Delta z}f_y\,\beta_y \;+\; p_y
    \end{bmatrix}
\end{equation}

\begin{equation}
    \text{If } \lambda \gg \Delta z 
    \implies
    \mathbf{\hat{x}} 
    \;\approx\;
    \begin{bmatrix}
        u \;+\; f_x\,\beta_x \\
        v \;+\; f_y\,\beta_y
    \end{bmatrix}
    \;=\;
    \begin{bmatrix}
        \hat{u} \\
        \hat{v}
    \end{bmatrix}
    \iff \mathbf{\hat{x}} \;\neq\; f(\lambda)
\end{equation}

This shows that the resulting transformed point is independent of the depth when the change in depth is small or when the depth itself is large. This approximation is justified when the depth is sampled densely in a dataset; in this case, changes in depth between nearby views are negligible. For nearest neighbors, this is the case for our datasets.

Lastly, we repeat the same procedure for any relative rotation about the out-of-plane camera $x$ and $y$ axes, which we expect to depend on precise depth information. For this example, we choose the $x$ axis, but the procedure is equivalent for the $y$ axis or both axes simultaneously.

\begin{equation}
    \mathbf{\hat{p}_C} \;=\; 
    R \ \mathbf{p_C} \;=\;
    \begin{bmatrix}
        1 & 0 & 0 \\
        0 & \cos\alpha & -\sin\alpha \\
        0 & \sin\alpha & \cos\alpha
    \end{bmatrix}
    \,\lambda 
    \begin{bmatrix}
        \tfrac{u}{f_x} \;-\; \tfrac{p_x}{f_x} \\
        \tfrac{v}{f_y} \;-\; \tfrac{p_y}{f_y} \\
        1
    \end{bmatrix}
    \;=\;
    \begin{bmatrix}
        \tfrac{\lambda}{f_x}\,(u - p_x) \\
        \cos\alpha\,\tfrac{\lambda}{f_y}\,(v - p_y) \;-\;\lambda\,\sin\alpha \\
        \sin\alpha\,\tfrac{\lambda}{f_y}\,(v - p_y) \;+\;\lambda\,\cos\alpha
    \end{bmatrix}
\end{equation}

\begin{equation}
    \mathbf{\hat{x}_{hom}} 
    = K \,\mathbf{\hat{p}_C}
    \;=\;
    \begin{bmatrix}
        f_x & 0   & p_x \\
        0   & f_y & p_y \\
        0   & 0   & 1
    \end{bmatrix}
    \begin{bmatrix}
        \tfrac{\lambda}{f_x}\,(u - p_x) \\
        \cos\alpha\,\tfrac{\lambda}{f_y}\,(v - p_y) \;-\;\lambda\,\sin\alpha \\
        \sin\alpha\,\tfrac{\lambda}{f_y}\,(v - p_y) \;+\;\lambda\,\cos\alpha
    \end{bmatrix}
    \iff \underline{\mathbf{\hat{x}} \;=\; f(\lambda)}
\end{equation}

As expected, when the homogeneous image point $\mathbf{\hat{x}_{hom}}$ is divided by its $z$ component to get the transformed image point $\mathbf{\hat{x}}$, it depends non-linearly on $\lambda$ which means that this is difficult to represent using VS as the transformed image point depends on precise depth information.

This shows that the C-L2 metric is an inappropriate choice as a distance norm for VS because of its isotropic behavior with respect to the x,y and z axes and its ignorance of isolated relative rotation. An approach to solving this is to take the sum of the normalized C-L2 metric together with the magnitude of the relative attitude, this was used for evaluation by the ESA-ACT + SLAB SPEC competitions \cite{Kisantal2020} ("$e_{pose}$") and authors who followed \cite{wang2022bridgingdomaingapsatellite,Park_2024multitaskacrossdomaingap,Chen2019,Black2021}.

This combined metric has the problem that the relative weights of position and attitude errors must be chosen manually. In general, the magnitudes of position and attitude are both real numbers but they have different dimensions (distances vs. angles); simply adding them is imprecise. In addition, attitude is bounded and can therefore easily be normalized while position is potentially unbounded. The BDD resolves these issues by being defined solely in the attitude space $S^3$ and considering only the rotations that "matter", namely the out-of-plane axes of rotation.

\subsection{Definition of the Boresight Deviation Distance}
\label{sec:bdddefinition}
\begin{align}
    BDD: S^{3}\times S^{3} \rightarrow \mathbb{R}, \quad BDD(q_1,q_2) &= \frac{|\theta|}{\pi} \left(1 - \left| \frac{2|\phi|}{\pi} - 1 \right|\right), \quad
    \phi \in (-\frac{\pi}{2},\frac{\pi}{2}], \, \theta \in(-\pi,\pi], \, BDD \in [0,1] \\
    \mathrm{where} \quad q_1 &\sim -q_1, \ q_2 \sim -q_2 \quad (S^3 \to SO(3)) \label{eq:identified}  \\
    \quad q_r &= r_r \ + \mathbf{v_r} = [r_r,q_{rx},q_{ry},q_{rz}] = q_1q_2^* \\
     \theta &= 2 \cdot \mathrm{atan2}(|\mathbf{v_r}|, r_r), \\
    \boldsymbol{\omega_r} &= [\omega_x,\omega_y,\omega_z] =  \frac{\mathbf{v_r}}{\mathrm{sin}(\frac{\theta}{2})} \ \mathrm{if} \ \theta \neq 0 \ \mathrm{else} \ \boldsymbol{\omega_r} = \mathbf{0} \\
    \boldsymbol{\omega_{r+}} &= [\omega_x,\omega_y,|\omega_z|] \quad \mathbf{e_z} = [0,0,1] \\
    \phi &= \mathrm{atan2}(|\mathbf{e_z} \times \boldsymbol{\omega_{r+}}|,\mathbf{e_z}^T\boldsymbol{\omega_{r+}})
\end{align}

The input to the BDD consists of two poses. First, the relative rotation $q_r$ between the attitude components is computed. This is converted to an axis angle representation, of which $\theta$ is the angular magnitude. The rotation axis of the relative rotation is obtained as $\boldsymbol{\omega_r}$ and mapped to the semi-sphere pointing towards the camera, resulting in $\boldsymbol{\omega_{r+}}$. Lastly, $\phi$ is the angle between $\boldsymbol{\omega_{r+}}$ and the camera boresight axis $\mathbf{e}_z$, computed with the numerically stable $\mathrm{atan2}(\dots)$.

\begin{figure}[t]
  \centering
  \begin{minipage}[t]{0.6\linewidth}
    \centering
    \includegraphics[width=\textwidth]{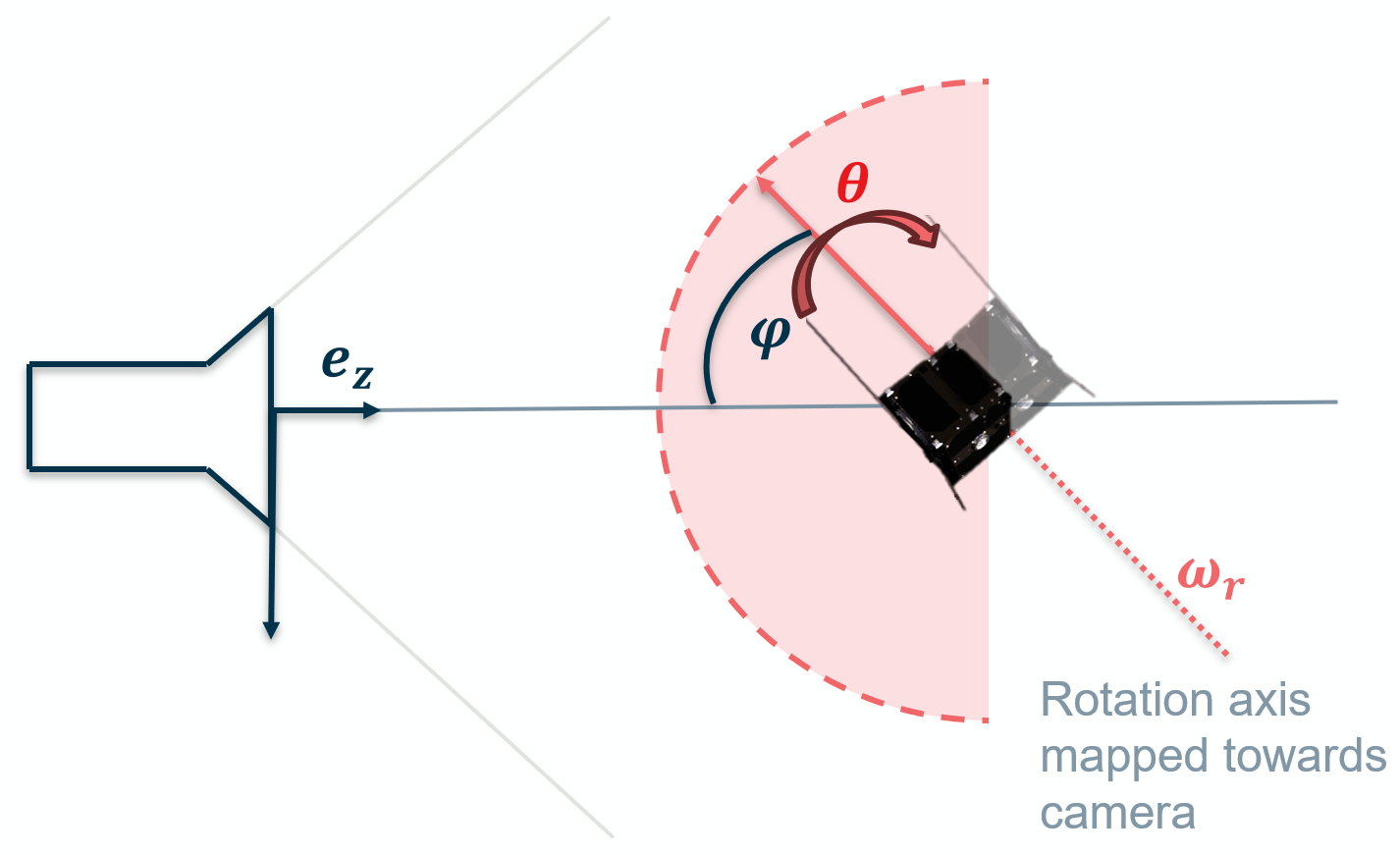}
    \caption{Illustration of the angles $\phi$ and $\theta$ of the BDD.}
    \label{fig:bdd2D}
  \end{minipage}%
  \hfill
  \begin{minipage}[t]{0.38\linewidth}
    \centering
    \includegraphics[width=\textwidth]{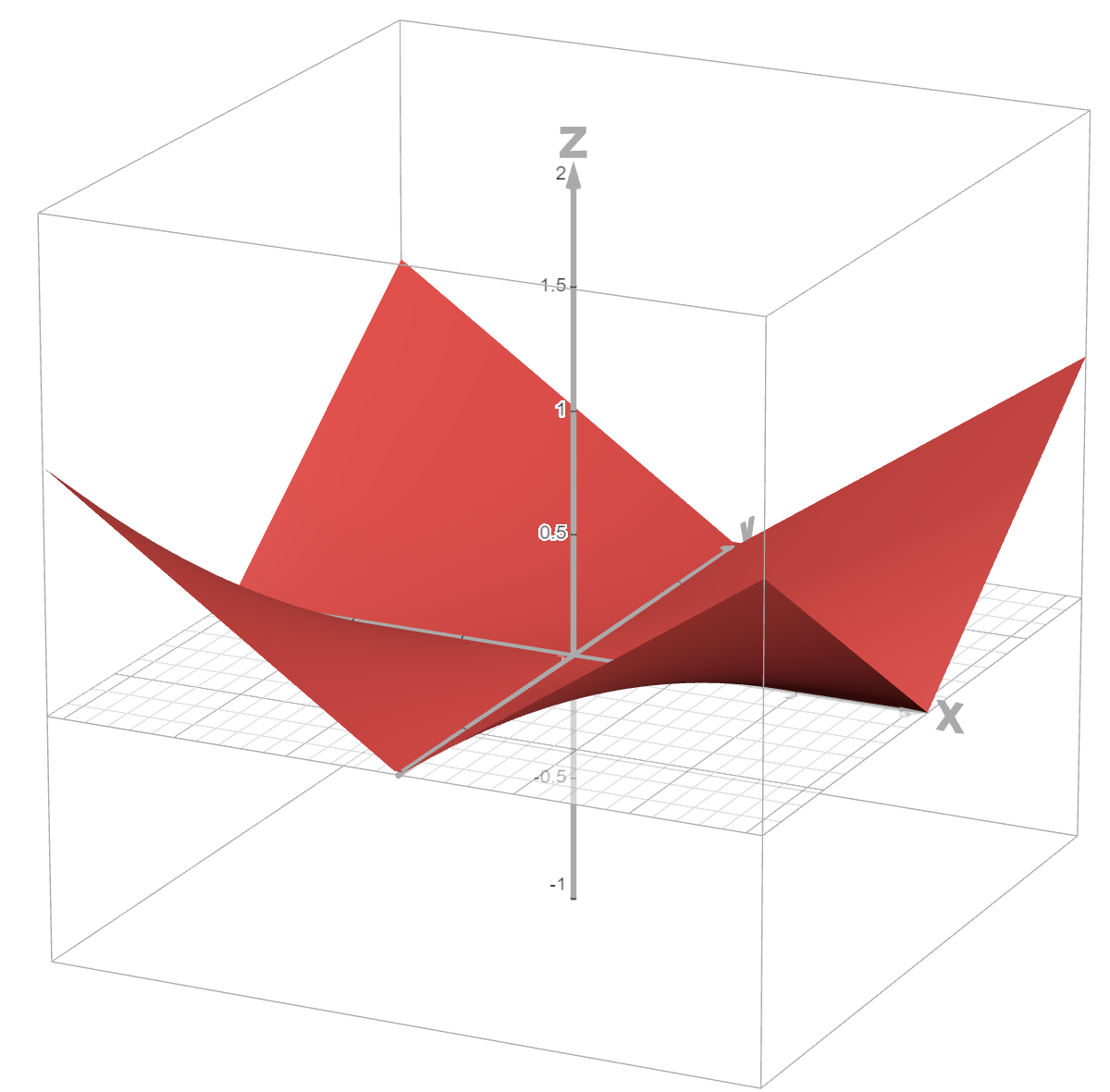}
    \caption{3D plot of the BDD, $x=\phi \in (-\frac{\pi}{2},\frac{\pi}{2}),\ y=\theta \in (-\pi,\pi), z = BDD$.}
    \label{fig:bdd3D}
  \end{minipage}
\end{figure}

The BDD is intuitively understood by considering the illustration in~\ref{fig:bdd2D}. In essence, the BDD encodes how far the target's rotation axis is from the boresight axis. To this end, the angle $\phi$ between the target's relative rotation axis and the boresight axis is combined with the magnitude $\theta$ of that rotation. Thus, the BDD vanishes when either of them is zero as both of these extremes can accurately be recreated using VS. This captures the previously shown anisotropy.

\subsection{Advantages of the Boresight Deviation Distance}
\label{sec:bddadvantages}
The advantages of the BDD, particularly with respect to C-L2, are the following: 
\begin{enumerate}
    \item \label{bdd:normalized} It is inherently normalized between 0 and 1, while C-L2 is potentially unbounded and needs manual tuning (w.r.t. the experimental setup, like shown by~\cite{park2024bridgingdomaingapflightready}) to normalize. This is advantageous for defining a density metric as simply the inverse of the BDD. This property also makes it dataset-agnostic: a given BDD value has the same meaning across different datasets without any manual tuning or thresholding necessary.
    \item It is anisotropic w.r.t.\ the six degrees of freedom of camera poses and only keeps track of the most important components. In contrast, C-L2 does not take into account the relative attitudes between camera poses. For C-L2, one must manually add a rotational term which degrades the properties of the resulting metric.
    \item \label{subsec:upperbound} It offers an upper bound for VS. Any target with a BDD > $0.5$ w.r.t. the source is not possible to synthesize. This is due to the fact that at greater values previously unseen faces of the target geometry always become visible.
    \item Lastly, the BDD fulfills the necessary conditions for a metric between members of the rotation group $SO(3)$: 
    \begin{align}
        (1) &\quad BDD(R,R) = 0 \\
        (2) &\quad R_1 \neq R_2 \Longrightarrow BDD(R_1,R_2) > 0 \\
        (3) &\quad BDD(R_1,R_2) = BDD(R_2,R_1) \\
        (4) &\quad BDD(R_1,R_3) \leq BDD(R_1,R_2) + BDD(R_2,R_3)
    \end{align}
    Where $R$ is a member of $SO(3)$. Due to the double cover of the unit quaternions, the antipodal points of $S^3$ (negation of a given unit quaternion $q \to -q$) must be identified by the map $q \to R$, as shown in~\ref{eq:identified}. In practice, this is achieved by ensuring that the scalar part of the relative quaternion $q_r$ is always positive.
\end{enumerate}

\vspace*{-0.5 cm}

\subsection{Density Measurement using the BDD}
\label{sec:Density}
One of the possible applications of VS is "densifying" existing datasets after they have been acquired or even planning an acquisition beforehand based on the desired density. To achieve this, a density metric must be established which serves to measure how densely the set of all poses is covered by a dataset.

Thanks to property~\ref{bdd:normalized} the BDD can be conveniently used to define this density metric. The density of a set of poses can be defined as the inverse of the size of the largest empty BDD-ball which fits into the set of poses. A BDD-ball is defined as $\{R: BDD(R_\text{query},R) < d\}$ where $d$ is the size and $R_\text{query}$ the center of the ball. The density is then: $\rho_{BDD} := \frac{1} {BDD_{max}}$ where $BDD_{max}$ is the size of the largest ball which does not contain any poses of a dataset ("empty"). $BDD_{max}$ is referred to in the following as \textbf{LB-BDD}. This measure can be spatially visualized as a cone, see~\ref{fig:bdd02cone}.

Using the LB-BDD means that the density of a dataset is not evaluated on a nearest-neighbor basis only. Doing so would unfairly bias in favor of trajectory-type datasets which are dense in the space of the trajectory but sparse in the whole space. The LB-BDD avoids this issue. In practice, to compute the LB-BDD, an ideal dense baseline sampling must be defined. We generate this sampling with a blue-noise-like pose generation algorithm inspired by "Mitchell's Best Candidate Algorithm"\cite{Mitchell1991}. The LB-BDD of a set of poses is then computed with respect to this ideal sampling using a min-max greedy nearest-neighbor algorithm.

\section{View Synthesis Methods}
In this chapter, we describe the general principles of view synthesis (\ref{VS_general}) and the two methods we chose to synthesize novel views (\ref{sec:Homography} and \ref{3DTransform}). 

\subsection{View Synthesis}
\label{VS_general}
VS renders novel 2D views from an existing dataset. Crucially, VS does not render, it generates new images from existing ones. Therefore, this method cannot deliver additional information beyond what is already present in the dataset, barring any optional data augmentation. What VS instead offers is to exploit the available data more fully.  

In general, to synthesize novel views we transform their nearest-neighbor (in terms of the BDD) that already belongs to the dataset to the queried novel pose. The justification for choosing the BDD for this purpose is shown formally in~\ref{sec:bddproof} and is also shown empirically in~\ref{table:correlations}. The nearest-neighbor (NN) search is performed by building a k-d-tree using the BDD, the tree can then quickly be queried to find the existing NN of a desired novel pose.

\begin{figure}[t]
    \makebox[\textwidth][c]{%
    \centering
    \begin{minipage}[t]{0.57\linewidth}
        \centering
        \includegraphics[width=\linewidth]{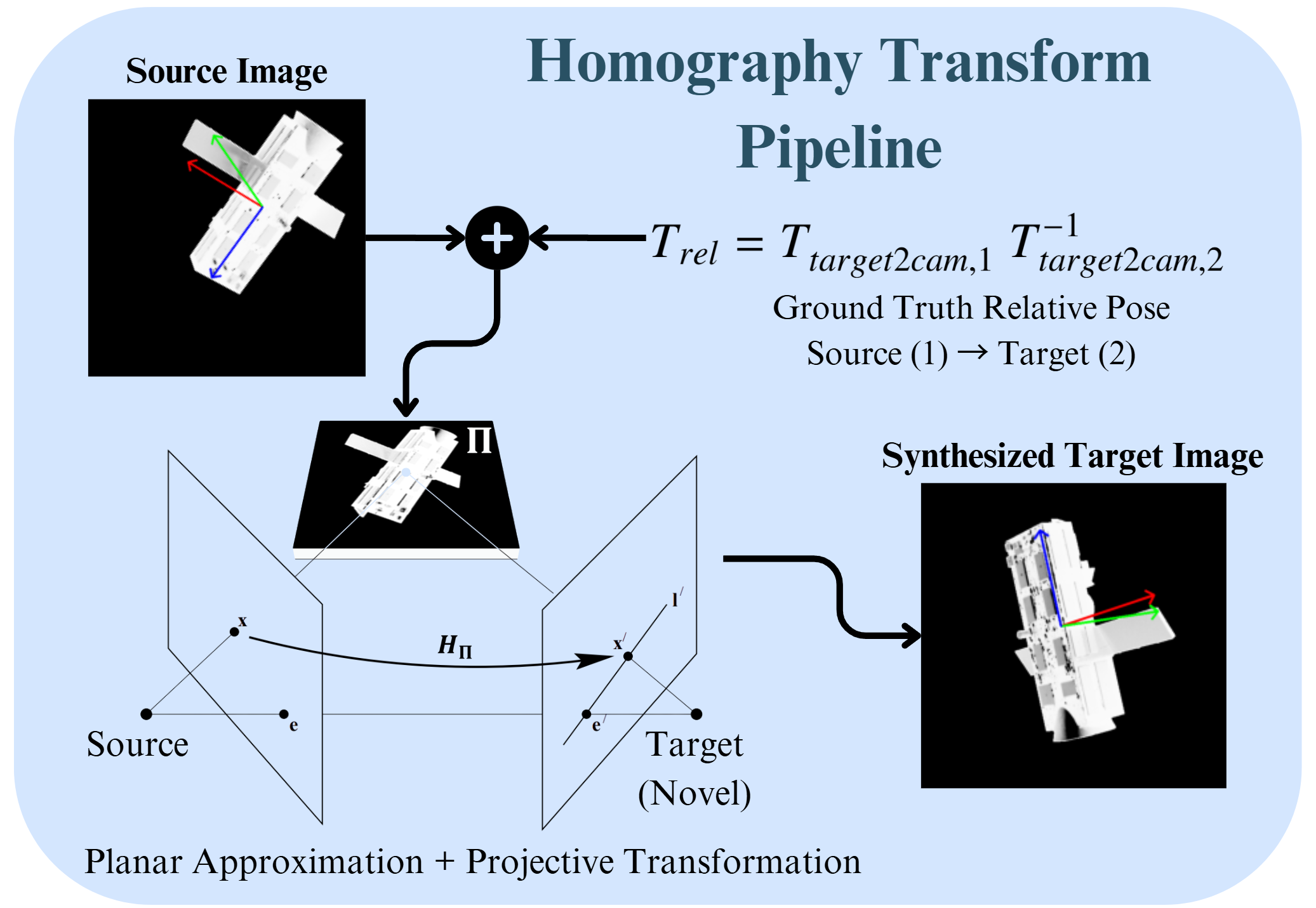}
        \caption{Principle behind the Homography Transform. Homography figure adapted from \cite{Hartley2013multipleviewgeometry}.}
        \label{fig:transform2dpipeline}
    \end{minipage}    
    \hfill
    \begin{minipage}[t]{0.57\linewidth}
        \centering
        \includegraphics[width=\linewidth]{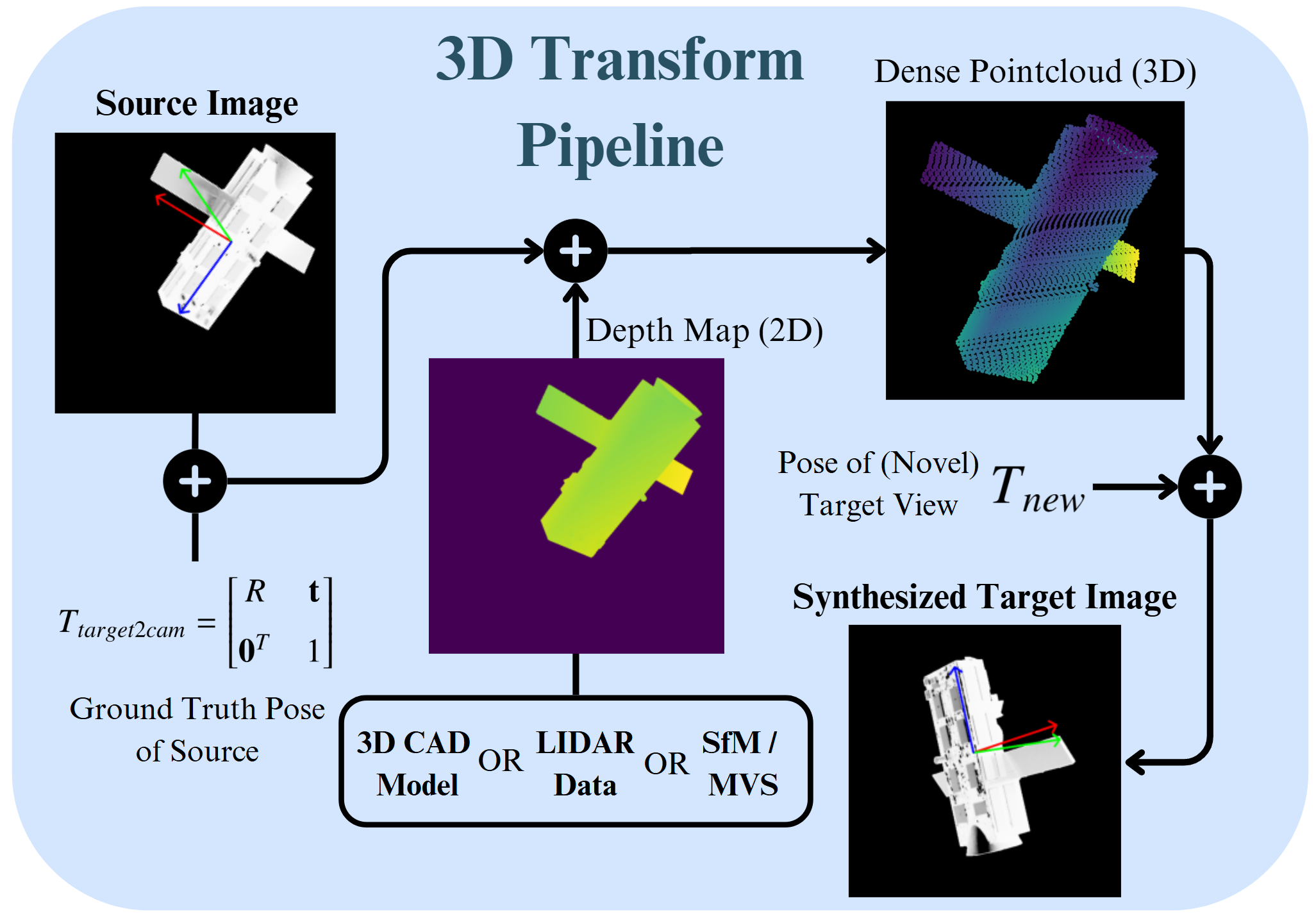}
        \caption{Principle behind the 3D Transform.}
        \label{fig:transform3dpipeline}
    \end{minipage}
    }    
\end{figure}

In this work, we use exclusively analytic, as in non-generative, methods for VS. The two proposed methods are named Homography and 3D Transform. A third one based on View Morphing \cite{Seitz1995} has been tested but needs further improvements in accuracy and runtime to be viable. This will be done in future work.  An important note concerning our VS methods is that they are not able to synthesize lighting changes between views, only geometric ones.

\subsection{Homography Transform}
\label{sec:Homography}
The first method is called the Homography Transform. It is based on the projective image transformation of the same name. It is the faster, but less accurate, of the two VS methods presented in this work.

The algorithm for the Homography Transform is described in~\ref{alg:homography} and is adopted from \cite{Hartley2013multipleviewgeometry}. The principle of this method is based on an approximation of the image as the projection of a planar scene. In our implementation, the depth to the corresponding world points is assumed to be constant, i.e.\ the normal vector of the plane is assumed to be parallel to the boresight axis. Note that the plane in~\ref{fig:transform2dpipeline} is inclined only for visual purposes; this is not done in the implementation.

\begin{algorithm}
\caption{Homography Transform}
\label{alg:homography}
\begin{algorithmic}[1]
\REQUIRE Source image $I_S$, source pose $T_S$, intrinsics $K$, optionally source mask $M$
\REQUIRE Target view's desired pose $T_T$
\ENSURE Synthesized target image $I_{T,\text{synth}}$

\STATE \textbf{Plane \& distance} \;\; $\mathbf{n} \leftarrow (0,0,1)^{\!T}, \quad d \leftarrow \lVert\mathbf{t_S}\rVert$
\STATE \textbf{Relative pose} \;\; $D \leftarrow \,T_T\;T_S^{-1}$  
\qquad $R_{12}\leftarrow D_{1:3,1:3}, \; \mathbf{t}_{12}\leftarrow D_{1:3,4}$
\STATE \textbf{World homography} \;\; $H \leftarrow R_{12} + \dfrac{\mathbf{t}_{12}\,\mathbf{n}^{\!T}}{d}$
\STATE \textbf{Image–plane map} \;\; $G \leftarrow K\,H\,K^{-1}$
\STATE \textbf{Warp} \;\; $I_{T,\text{synth}} \leftarrow \text{warpPerspective}(I_S, G)$
\STATE \textbf{Transformed mask} $M' \leftarrow \text{warpPerspective}(M, G,\text{border}=0)$
\RETURN $I_{T,\text{synth}}$, optionally $M'$
\end{algorithmic}
\end{algorithm}

We use the \texttt{warpPerspective} routine from \texttt{OpenCV}, which executes a perspective transformation. Optionally, the algorithm allows for providing the mask and 2D keypoint locations for the input image to obtain their transformed counterparts. We use the transformed mask to evaluate the accuracy of the synthesized view.

\subsection{3D Transform}
\label{3DTransform}
The 3D Transform addresses the key limitation of the Homography Transform, that being its inaccurate depth approximation, by leveraging available depth data (e.g., a depth-map rendered with a 3D CAD model or LIDAR capture). If neither is available, the 3D Transform cannot be used; in that case, the Homography Transform shall be used. If the latter delivers unsatisfactory results a multi-reference method like View Morphing shall be used. 

Unlike the Homography Transform’s single perspective warp, the 3D Transform operates on each pixel individually, which may leave gaps in the target view. Strategies for filling these gaps are discussed below. In our implementation, depth data comes from a depth-map rendered using the 3D CAD model of the target. The rendering of the depth-map in our implementation is done with \texttt{pyrender}'s \texttt{OffscreenRenderer}. The grayscale intensity value of each pixel in the depth-map gives the $z$-coordinate in the camera reference frame to the underlying 3D world point of the same pixel in the image; the 3D point is then found by evaluating the back-projected camera ray at that depth value.

Because pixels are handled one-by-one, two distinct 3D points can project to the same target pixel, creating “intra-image” gaps. (Extra-image gaps i.e. out-of-focus-areas~\cite{DeOliveira2021} are resolved by masking the input image). Intra-image gaps are filled using morphological operations and interpolation: a sliding $5\times5$ convolution window gathers valid neighbors (detected via a mask that saves where source pixels land in the target image) and computes color values for the gaps.

See \ref{alg:3d_transform} for a detailed description of the 3D Transform's algorithm.
 
\begin{algorithm}
\caption{3D Transform with interpolation}
\label{alg:3d_transform}
\begin{algorithmic}[1]
\REQUIRE Source image $I_S$, source pose $T_S$, intrinsics $K$, 3D model of target object $\mathcal{O}$
\REQUIRE Target view's desired pose $T_T$
\ENSURE Synthesized target image $I_{T,\text{synth}}$

\STATE Render source depth map $D_S$ using $\mathcal{O}$, $K$ and $T_S$
\FOR{each pixel $\tilde{p}=(u,v,1)$ in $I_S$}
    \STATE \textbf{Camera coordinates} \;\; $P_c \leftarrow D_S(K^{-1}\,\tilde{p})$ \COMMENT{evaluate ray at depth map pixel value}
    \STATE \textbf{World coordinates} \;\; $P \leftarrow R_S^{\top}(P_c - \mathbf{t_S})$
    \STATE \textbf{Target image coordinates} \;\; $\tilde{p'} \leftarrow K \,(R_TP+\mathbf{t_T})$
    \STATE $I_{T,\text{synth}}(\tilde{p'}) \gets I_S(\tilde{p})$
\ENDFOR
\STATE \textbf{Valid mask} \;\; $V_T=\mathbf{-1}$, override with transformed pixels
\FOR{each gap $h \in \{\,h \mid V_T(h)=-1\,\land D_S(h)\geq0\}$}
    \STATE Interpolate $I_{T,\text{synth}}(h)$ from neighboring valid pixels in a $5\times5$ window
\ENDFOR
\STATE \textbf{return} $I_{T,\text{synth}}$
\end{algorithmic}
\end{algorithm}

One possibility to improve the algorithm is to not detect gaps only based on the transformed coordinates of the source pixels but to fill all the pixels belonging to the target model in the synthesized image. These pixels could be detected using the target image's mask of the target model which can easily be obtained by thresholding the target image's depth-map. In this way, disocclusions that arise from new faces of the target model becoming visible, which are ignored in the above algorithm, would also be filled. However, this is not done this in this work to not bias the results of the IoU metric (which would always be 1 otherwise, as the whole mask would be filled).

\vspace*{-0.25cm}
\section{Validation}
\label{sec:headervalidation}
To validate VISY-REVE, we create a general performance model for VS by means of a Monte Carlo (MC) campaign in which, for each draw, a synthesized image is compared with the actual existing image with the same pose. The MC results are then used to derive performance specifications that show the limits of VS and that can inform its real-world use. To this end, first, the necessary quality metrics to evaluate VS will be defined (\ref{sec:qualitymetrics}). Then, the MC process is explained in detail (\ref{subsec:implementation}). The results of MC are described in~\ref{sec:results}. The practical application performance of VS will be analyzed through two test-cases which model possible real-life use-cases. Their results are described in~\ref{sec:testcase1,traj} and~\ref{sec:testcase2,densification}.  

\subsection{Quality Metrics}
\label{sec:qualitymetrics}
Two different types of metrics are needed to evaluate VS. The first comprises quality metrics that evaluate how similar a synthesized image is to the actual image. The second includes distance metrics that measure the distance between poses; these should also be predictive of the accuracy of VS, as explained in~\ref{sec:bddmotivation}. Together, they can be used to derive the performance model. As a distance metric, the BDD is chosen, which is designed for this purpose. 

As quality metrics, multiple ones are chosen to cover different aspects of VS accuracy. The chosen metrics are all full-reference, i.e.\ they compare a synthesized image with its ideal counterpart. We opt for a balance between optical, i.e. metrics that identify radiometric differences; structural, i.e. metrics that identify geometric differences; and functional, i.e. metrics that represent the functioning of VBN algorithms. The chosen metrics are the following:

\begin{enumerate}[leftmargin=1.5em,itemsep=0.5pt]
    \item Optical, \textbf{SSIM}. The Structural Similarity metric \cite{Wang2004}. The SSIM is computed using images cropped to the bounding boxes of the target model to remove bias w.r.t. its size. 
    \item Structural, \textbf{IoU}. The Intersection over Union between two masks of the target model: The transformed source mask for the synthesized view $M'$ and the mask for the actual view $M_T$, formally: \(IoU(M', M_T) = \frac{M' \cap M_T}{M' \cup M_T}\).
    \item Functional, \textbf{KPS-L2},  Mean of the Euclidean reprojection errors between ground-truth 2D keypoints $\mathbf{u}$ and as detected by a convolutional neural network (CNN) $\hat{\mathbf{u}}$. Keypoints are landmarks defined on the 3D geometry of the target model that can be projected into each image based on the ground-truth pose. We opt for detecting keypoints in the synthesized images with a CNN instead of directly using their transformed coordinates as the CNN detections also encode the visual fidelity of the synthesized images, not just their geometric accuracy. Formally: \(KPS\text{-}L2 = \frac{1}{K} \sum_{i=1}^{K} \left\| \hat{\mathbf{u}}_i - \mathbf{u}_i \right\|_2\) where $K$ is the number of keypoints. 
    \item \label{item:kpsvbn} Functional, \textbf{KPS-VBN}. Ratio between: 1. The mean of the Euclidean distances between the ground-truth 3D keypoints $\mathbf{X}$ and the back-projected CNN-detected 2D keypoints $\hat{\mathbf{X}}$. The back-projection is done using the ground-truth pose. This is divided by: 2. The Euclidean distance between the camera and the target model, resulting in a relative range metric. This kind of metric is typical for VBN missions, a common baseline is $1\%$. It makes the error in the keypoints invariant to the range, as shown by~\cite{Hu2021}. Formally: \(KPS\text{-}VBN = \frac{\frac{1}{K} \sum_{i=1}^{K} \left\| \hat{\mathbf{X}}_i - \mathbf{X}_i \right\|_2}{\left\| \mathbf{t} \right\|_2}\).
\end{enumerate}

The neural network employed for metrics 3. and 4.\ uses the HRNet architecture~\cite{wang2020deephighresolutionrepresentationlearning} which is state-of-the-art for pose estimation by keypoint detection in VBN~\cite{CassinisPHD2022,wang2022bridgingdomaingapsatellite}. It is trained to regress heatmaps for the 2D projections of the 3D keypoints. The number of keypoints chosen is 8 for SWISSCUBE, 9 for SPEED+ and 16 for ESA-AIRBUS, their positions are chosen manually to coincide with salient features on the target models. The network contains around $63.5$ million parameters. For training and inference, the input images are normalized and resized to $288\times384$ pixels with an output heatmap size for each keypoint of $144\times192$. The network was trained for 20 epochs with a learning rate of $0.0005$, achieving a final PCK test score (percentage of correct keypoints) of at least 0.995 for all datasets, which means that $99.5\%$ of detected keypoints fall within $5\%$ of the heatmap width w.r.t. their ground truth locations.

\subsection{Implementation}
\label{subsec:implementation}

The MC process uses a strategy we name \textbf{sample replacement}, similar to \cite{Perez2024}. This involves continuously drawing new pairs of samples from a dataset. The first image of the pair, named \textit{target}, is synthesized from the second one, named \textit{source}. The synthesized and actual target images are then compared with the quality metrics described above. The MC is carried out with a sample size of $10.000$ sample replacements for each dataset. The pairs are chosen randomly, albeit so that their BDD is below $0.5$, see~\ref{subsec:upperbound}. The background is masked out on all source images to reduce bias. 

\begin{figure}[t]
  \centering
  \makebox[\textwidth][c]{%
    \begin{minipage}[t]{0.5\linewidth}
      \centering
      \includegraphics[width=0.62\linewidth]{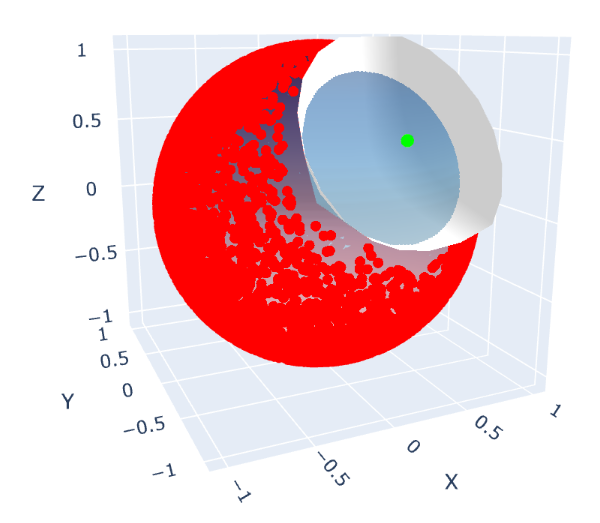}
      \captionof{figure}{A LB-BDD of 0.2 visualized as a white cone. Each red point marks a camera position, normalized to the unit sphere, ball center marked in green (technically a line).}
      \label{fig:bdd02cone}
    \end{minipage}\hfill
    \hspace*{0.2cm}
    \begin{minipage}[t]{0.3\linewidth}
      \centering
      \includegraphics[width=\linewidth]{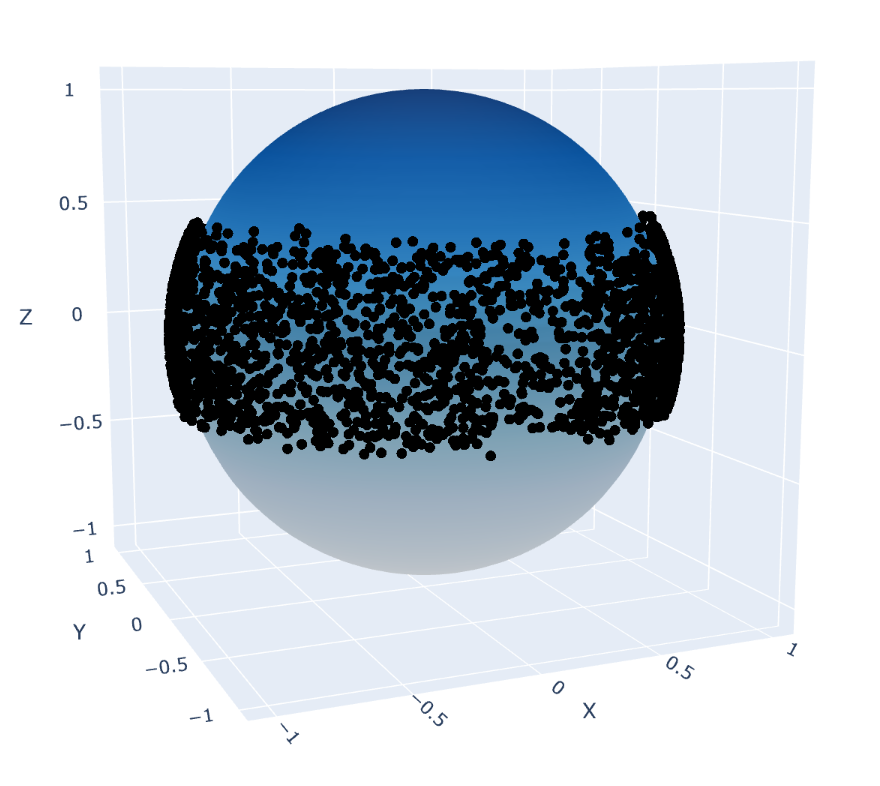}
      \captionof{figure}{Initial ESA-AIRBUS dataset with missing poles. Camera positions normalized to unit sphere.}
      \label{fig:initialESAAirbus}
    \end{minipage}
    \hspace*{0.2cm}
    \begin{minipage}[t]{0.3\linewidth}
      \centering
      \includegraphics[width=\linewidth]{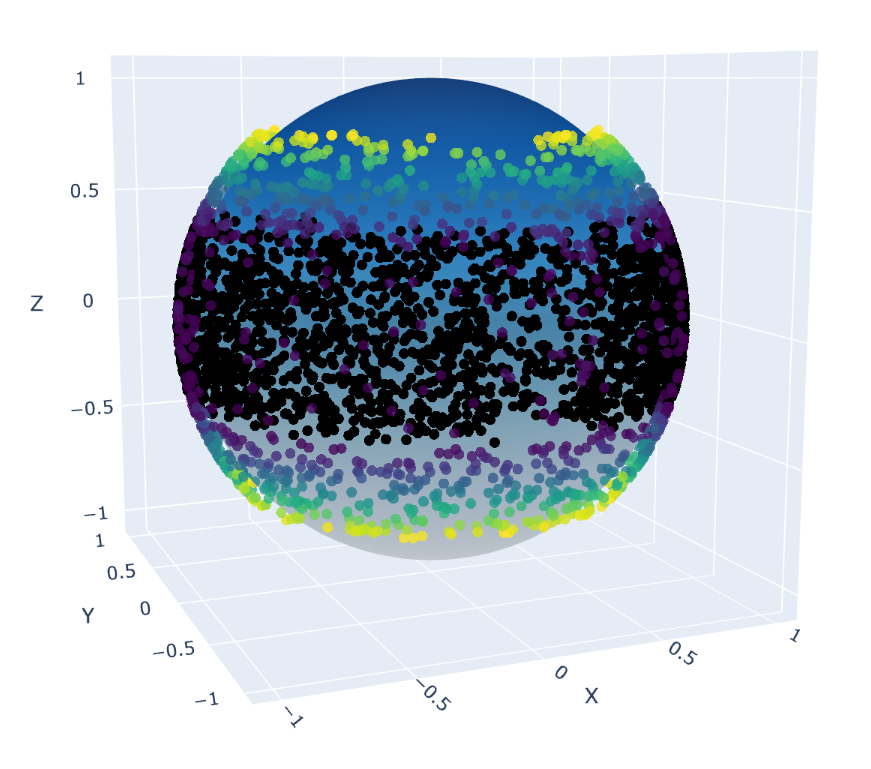}
      \captionof{figure}{Densified ESA-AIRBUS with new views up to a BDD of 0.1. Brighter means a higher BDD from source to target.}
      \label{fig:densifiedESAAirbus}
    \end{minipage}%
  }
  \label{fig:boxed-panels}
\end{figure}

\vspace*{-0.25cm}

\section{Results}
\label{sec:results}

\subsection{Performance Model}
\begin{figure}[t]
  \centering
  \makebox[\textwidth][c]{%
    \begin{minipage}[t]{0.4\linewidth}
      \centering
      \includegraphics[width=\linewidth]{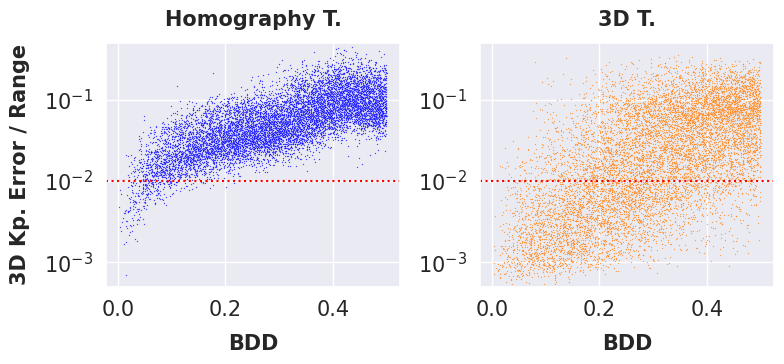}
      \subcaption{SPEED+}
      \label{fig:speedplus}
    \end{minipage}
    \begin{minipage}[t]{0.4\linewidth}
      \centering
      \includegraphics[width=\linewidth]{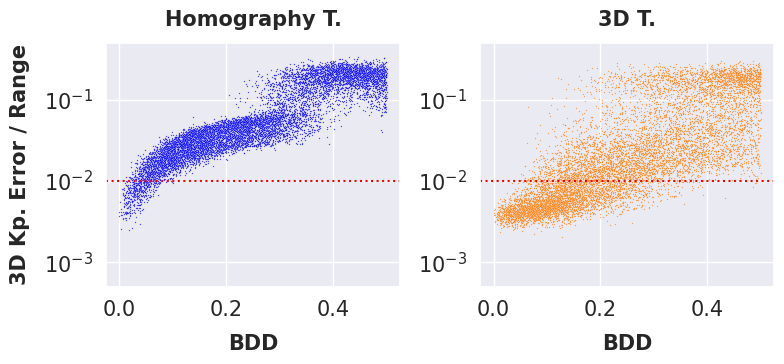}
      \subcaption{ESA-AIRBUS}
      \label{fig:airbusvbnv4}
    \end{minipage}
    \begin{minipage}[t]{0.4\linewidth}
      \centering
      \includegraphics[width=\linewidth]{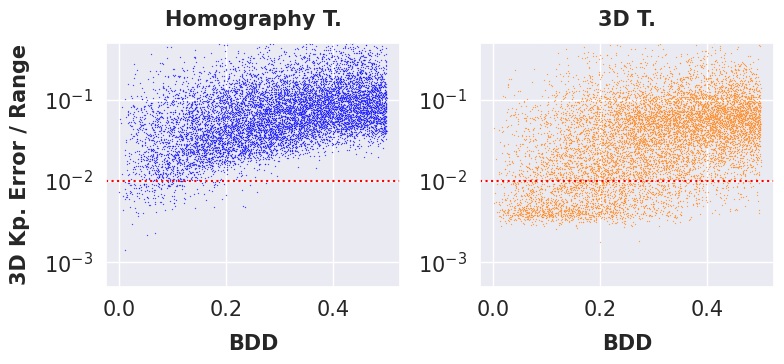}
      \subcaption{SWISSCUBE}
      \label{fig:swisscube}
    \end{minipage}%
  }
  \caption{MC results for the KPS-VBN quality metric ("3D Kp. Error / Range"). The red horizontal lines mark the $1\%$ VBN relative range-requirement. Note that the y axis is logarithmic, its limits and scale have been kept the same across all plots. Table~\ref{table:montecarlo} can be interpreted as vertical cutoff lines in these plots at the BDD values given in the table.}
  \label{fig:montecarlokpsvbn}
\end{figure}

The results of the MC are shown in detail in~\ref{table:montecarlo} and are visualized for the KPS-VBN metric in~\ref{fig:montecarlokpsvbn}. Table~\ref{table:montecarlo} shows the maximum LB-BDD the respective datasets could have so that VS would fulfill the quality requirements with a $3\sigma$ confidence. Using confidence intervals is necessary because of occasional outliers. As an example for the $1\%$ range requirement (KPS-VBN metric~\ref{item:kpsvbn}), using the best VS method, the 3D Transform, the results are: $0.087$ for ESA-AIRBUS, $0.024$ for SPEED+ and $0.001$ for SWISSCUBE (higher is better). The actual LB-BDD values of these datasets are $0.31$ for ESA-AIRBUS, $0.0045$ for SPEED+ and $0.0071$ for SWISSCUBE (lower is better), see~\ref{sec:Density}. 

This means that for SPEED+, since $0.024$ > $0.0045$, \textbf{every possible novel pose} can be synthesized and fulfill the $1\%$ range-requirement with greater than $99.7\%$ confidence. For ESA-AIRBUS and SWISSCUBE this is not the case, namely because ESA-AIRBUS has a very high LB-BDD with its missing poles as shown in~\ref{fig:initialESAAirbus} and for SWISSCUBE because of the small target, which is difficult to recognize by our CNN for even small deviations induced by VS.

\begin{table}[H]
\setlength{\tabcolsep}{12pt}
\makebox[\linewidth][l]{%
  \hspace*{1.8 cm}%
  \begin{tabular}{lcccccc}
    \hline
    \multirow{2}{*}{\textbf{Dataset}} &
    \multicolumn{2}{c}{\textbf{KPS-VBN} $3\sigma- 1\%$} &
    \multicolumn{2}{c}{\textbf{IoU} $3\sigma-0.9$} &
    \multicolumn{2}{c}{\textbf{SSIM} $3\sigma-0.9$} \\
    & \textbf{HOM} & \textbf{3DT} & \textbf{HOM} & \textbf{3DT} & \textbf{HOM} & \textbf{3DT} \\
    \hline
      ESA-AIRBUS &  0.044 & 0.087 & 0.0304 & 0.097 & -  & - \\
      SPEED+  & 0.016 & 0.024 & 0.0044 & 0.016 & 0.0041 & 0.0041 \\
      SWISSCUBE & 0.001 & 0.001 & 0.001 & 0.001 & 0.0025 & 0.001 \\
    \hline
  \end{tabular}
}
\caption{3-sigma MC results. The values in the cells are LB-BDD. For KPS-VBN we measure the maximum BDD below which $3\sigma\%$ of sample replacements fulfill the $1\%$ VBN range-requirement. For IoU and SSIM we measure the max. BDD below which $3\sigma\%$ of sample replacements are above 0.9. HOM is Homography, 3DT is 3D Transform. Higher is better. Values rounded to two significant figures. SSIM requirement for ESA-AIRBUS never fulfilled.}
\label{table:montecarlo}
\end{table}

\subsubsection{Correlations BDD vs. C-L2}
We empirically show the predictive capabilities of the BDD compared to C-L2 w.r.t.\ the accuracy of VS. For this comparison, we calculate the Pearson Correlation Coefficients \cite{freedman2007statistics} between the quality metrics (KPS-L2, IoU, SSIM) and the BDD / C-L2 for the MC sample replacement results. The correlations values are shown in~\ref{table:correlations}. The entries in bold mark the distance metric with the higher correlation. 

\begin{figure}[t]
    \centering
    \makebox[\textwidth][c]{%
        \includegraphics[width=1.1\linewidth]{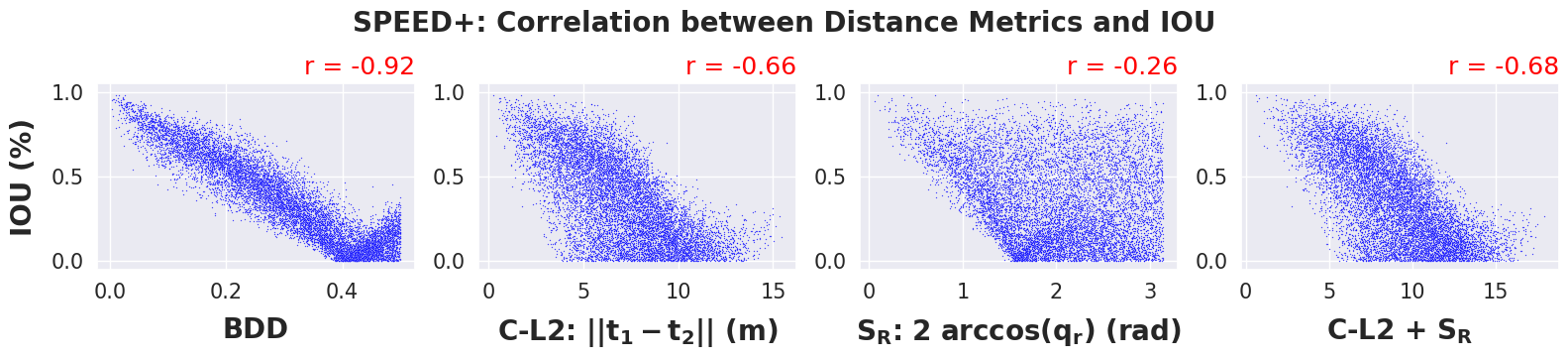}
    }
    \caption{Correlation results of the IoU metric for SPEED+. It can be clearly seen that the BDD better predicts the IoU than any other distance metric. Red numbers indicate Pearson Correlation Coefficients. Compare~\ref{table:correlations}.}
    \label{fig:correlationsmetricsfigs}
\end{figure}

The BDD generally outperforms C-L2 in the KPS-L2 quality metric, which best represents the current use-case of VBN algorithms. Only for the SSIM C-L2 is mostly superior. This is relativized by the fact that the correlations for the SSIM are generally lower and that for all metrics, when the C-L2 is better, the margins to the BDD are small. We can conclude that the BDD is indeed more representative of the mechanism of VS and that it is thus the better distance metric. In~\ref{fig:correlationsmetricsfigs} we visually show this for SPEED+, also compared with the rotation magnitude and SPEC metrics.

\begin{table}[ht]
\setlength{\tabcolsep}{12pt}
\makebox[\linewidth][l]{%
  \hspace*{0.45 cm}%
  \begin{tabular}{llcccccc}
    \hline
    \multirow{2}{*}{\textbf{Distance Metric}} &
    \multirow{2}{*}{\textbf{Dataset}} &
    \multicolumn{2}{c}{\textbf{KPS-L2}} &
    \multicolumn{2}{c}{\textbf{IoU}} &
    \multicolumn{2}{c}{\textbf{SSIM}} \\
    & & \textbf{HOM} & \textbf{3DT} & \textbf{HOM} & \textbf{3DT} & \textbf{HOM} & \textbf{3DT} \\
    \hline
    \multirow{3}{*}{\textbf{BDD}}
      & ESA-AIRBUS    & \textbf{0.83} & 0.62         & -0.94          & -0.76 & -0.37        & -0.35 \\
      & SPEED+    & \textbf{0.60} & \textbf{0.45} & \textbf{-0.92} & \textbf{-0.54} & -0.10 & \textbf{-0.27} \\
      & SWISSCUBE & \textbf{0.29} & \textbf{0.27} & \textbf{-0.58} & -0.34 & \textbf{0.08} & -0.08 \\
    \hline
    \multirow{3}{*}{\textbf{C-L2}}
      & AIRBUS    & 0.82 & 0.62 & -0.94         & \textbf{-0.77} & \textbf{-0.39} & \textbf{-0.36} \\
      & SPEED+    & 0.19 & 0.16 & -0.66         & -0.36         & \textbf{-0.12} & -0.25 \\
      & SWISSCUBE & 0.06 & 0.11 & -0.53         & \textbf{-0.35} & 0.02  & \textbf{-0.12} \\
    \hline
  \end{tabular}%
}
\caption{Correlation table BDD \& C-L2. HOM is Homography Transform, 3DT is 3D Transform. Higher is better.}
\label{table:correlations}
\end{table}

\vspace*{-0.1cm}

\subsection{Functional Validation}

\subsubsection{Test-case 1: Real-Time Trajectory Generation}
\label{sec:testcase1,traj}
One of the main use-cases of VISY-REVE is the real-time in-the-loop synthesis of images, only needing a sparse dataset as input. VISY-REVE can be used as a replacement for synthetic rendering or laboratory acquisition to close the loop for testing image processing algorithms. Combined with a GNC stack, VS therefore enables real-time \textbf{novel trajectory synthesis} \cite{wang2024freevsgenerativeviewsynthesis}. Note also that VS is even able to synthesize images that lie outside the convex hull of the dataset. For best results, however, the lighting position should remain the same or similar across the dataset. 

\begin{figure}[t]
  \centering
  \noindent
  \makebox[\textwidth][c]{%
    \begin{minipage}[t]{0.42\linewidth}
      \centering
      \includegraphics[width=\linewidth]{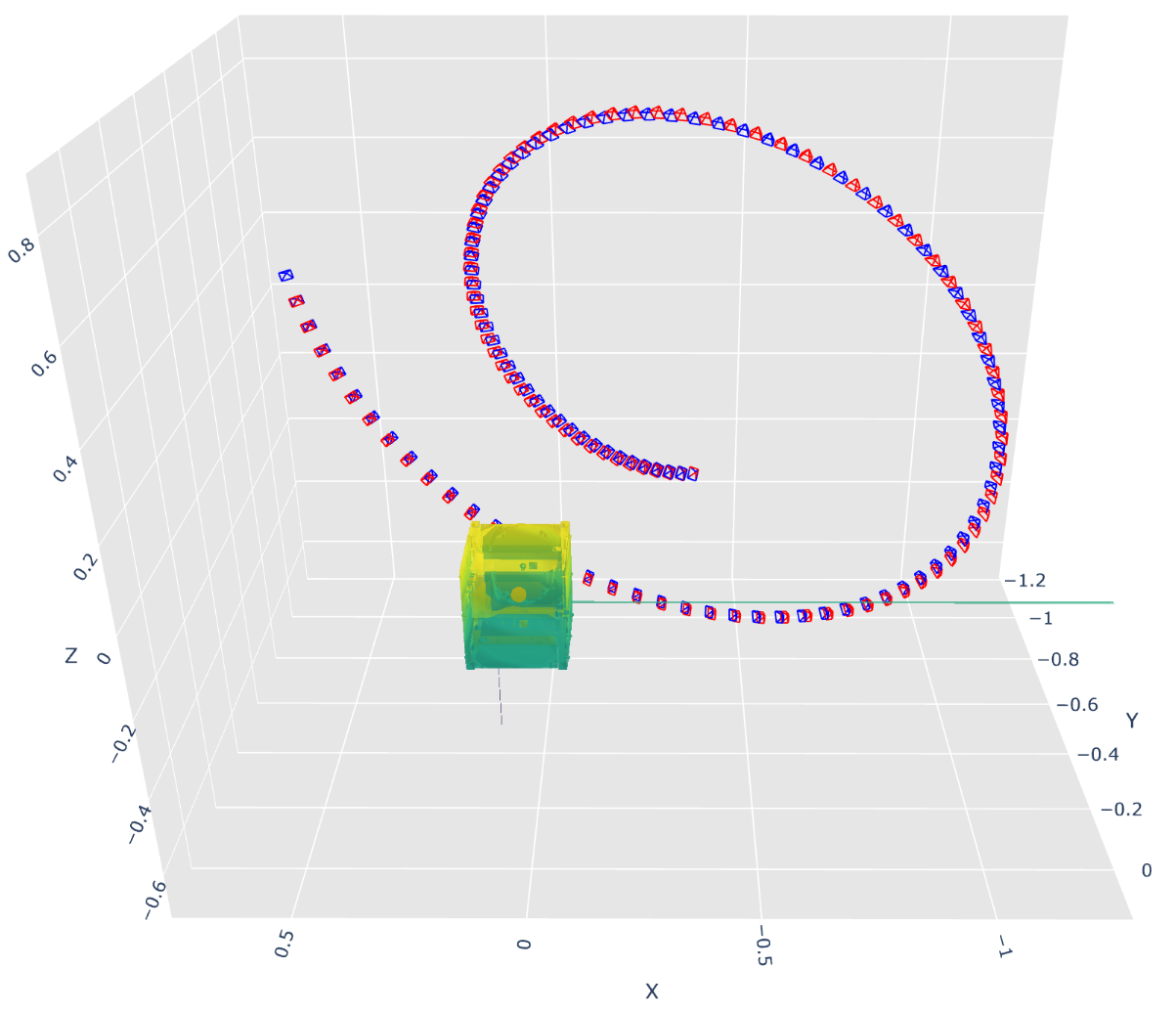}
      \caption{Existing trajectory (blue) and synthesized trajectory (red) from SWISSCUBE.}
      \label{fig:traj_mesh}
    \end{minipage}
    \hspace{1em}
    \begin{minipage}[t]{0.63\linewidth}
      \centering
      \includegraphics[width=\linewidth]{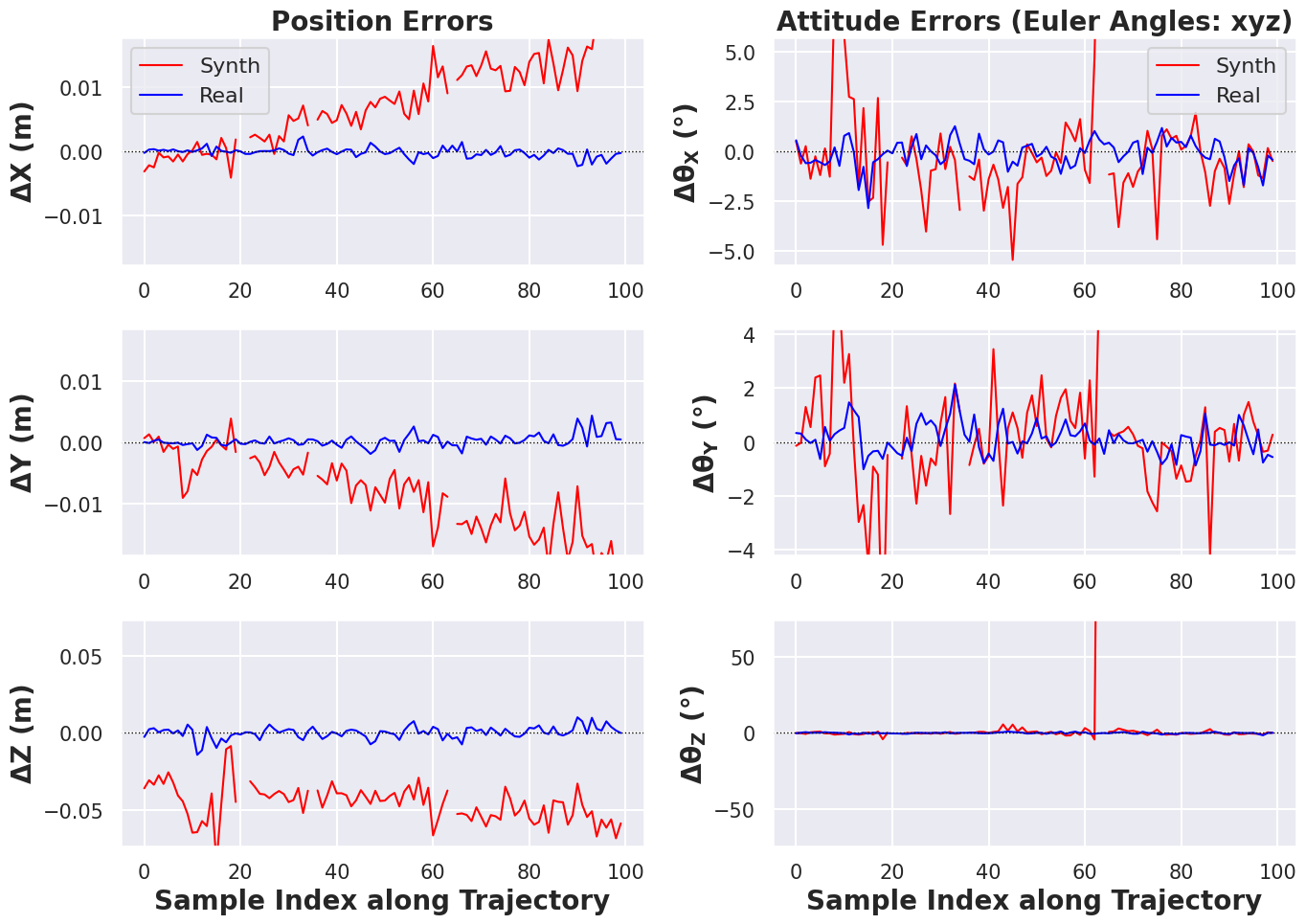}
      \caption{HRNet + PnP performance for trajectories shown to the left. Y-axes clipped to $3\sigma$ bounds of the combined errors.}
      \label{fig:pnp}
    \end{minipage}%
  }
\end{figure}

We demonstrate this use-case qualitatively: we synthesize a trajectory with small deviations from an existing one, see~\ref{fig:traj_mesh}. Pose estimation is done for both trajectories using the same CNN as before, combined with PnP. The errors are shown in~\ref{fig:pnp}. As can be seen, the poses can accurately be reconstructed from the synthesized images most of the time. 

\subsubsection{Test-case 2: Densifying Datasets}
\label{sec:testcase2,densification}
The second promising use-case of VISY-REVE is to \textit{densify}, i.e. increase the number of samples in a dataset in an optimal manner. To this end, first, regions of poor sampling in the dataset are identified using the BDD and the ideal sampling described in~\ref{sec:Density}. Then, VS is used to populate these regions with images and ground-truth labels.

The methodology of identifying under-sampled regions using the BDD can also be employed at dataset creation-time. In that case, the image poses to acquire can be defined w.r.t.\ the density requirements for accurate VS found in~\ref{table:montecarlo}. Thereby, R.T. or synthetic acquisition campaigns are shortened and gaps can be filled using VS on-line or off-line.

We demonstrate this use-case in~\ref{fig:densifiedESAAirbus}. Here, the ESA-AIRBUS dataset is densified using the 3D Transform and its density value $\rho_{BDD}$ is increased from 3.33 to 5 (LB-BDD: 0.3 to 0.2), an increase of $50\%$. This exceeds the $3\sigma$ limit (cf.~\ref{table:montecarlo}). However, for ESA-AIRBUS the $2\sigma$ KPS-VBN $1\%$ range-requirement with the 3D T. is fulfilled up to a LB-BDD of $0.11$. Therefore, we see that the densified dataset will fulfill the requirement more than $95\%$ of the time.

\vspace*{-0.25cm}

\subsubsection{Runtime Analysis}
The runtime of VS is analyzed. The tests are performed on a laptop with an AMD Ryzen 5 3500U CPU at 1400 clock speed, 16 GB RAM, AMD Radeon(TM) Vega 8 GPU. The algorithms are all run single-threaded. The benchmark is the synthesis of a random cubic spline trajectory around the origin of 10 m length containing 500 samples for the SPEED+ dataset with an image size of 1200x1920 pixels. The results are rounded to three significant figures.

The results for the Homography Transform is an 0.476s average, including rendering the depth-map (to mask the background on the input image) and including loading the image from disk. Without masking the input, so without depth-map rendering, this drops to 0.0119s on average. For the 3D Transform with interpolation, it takes 0.852s on average, including rendering the depth map (necessary to perform the transform and to mask the background) and including loading the image from disk. Without interpolation, this drops to 0.699s on average. 

\section{Discussion}
Through the MC campaign and the resulting performance model, it has been successfully shown that for SPEED+, VS has the necessary accuracy to synthesize \textbf{all possible novel poses} while fulfilling the $1\%$ range-requirement for the CNN-based keypoint detection error with greater than $99.7\%$ confidence. For the SWISSCUBE and ESA-AIRBUS datasets, not every novel pose can be synthesized at this level, although for both datasets, every NN can be synthesized to fulfill the requirement, as the average NN-BDD values for these datasets are near zero. 

The results of the two test-cases indicate that the presented VISY-REVE pipeline, namely dataset evaluation and VS, is useful for real-life validation of VBN image processing algorithms, although more quantitative experiments are needed to confirm this. Lastly, we have shown that the BDD is a superior metric for choosing NN views for VS as it better represents the drop in quality incurred by an increase in distance.

We acknowledge that having chosen a CNN as a validation technique may have negatively impacted the presented findings, as the results are sensitive to the quality of the trained network. In particular, there is a domain gap between synthesized and actual images, which is not present during training. This gap is due to deviations in the images induced by VS, namely deformations for the Homography Transform and gaps for the 3D Transform. Using data augmentation as shown by~\cite{Park_2024multitaskacrossdomaingap,CassinisPHD2022,park2024bridgingdomaingapflightready} may improve the findings as the CNN would learn to better generalize across domains.

The \textit{complete source code of our work is released publicly}; please see the project page for more information.

\section{Further Work}
Speed is a crucial factor for the adoption of VISY-REVE. For this work, all algorithms were implemented in Python for a single-threaded CPU runtime and were not yet sufficiently optimized. For better performance, the algorithms would have to be converted to C/C++. GPU acceleration is also viable, as the VS methods are easily parallelizable.

To further the development of the BDD, a closed-form solution for a desired rotation given a BDD and a query rotation should be developed which would be useful to avoid having to use rejection sampling as in this work. The BDD should also be enhanced to take into account foreshortening effects associated with translation of the camera.

Additionally, using more and potentially better VS methods is possible. In particular, using e.g.\ View Morphing which can combine multiple source views to yield the target view might result in better accuracy as this would enable synthesizing lighting differences associated to changes in pose. Furthermore, we plan to implement modern VS methods such as NeRF~\cite{mildenhall2020nerfrepresentingscenesneural} and Gaussian Splatting, which are also capable of synthesizing lighting changes.

Lastly, as mentioned before, the CNN used for validation should be trained with photometric and geometric data augmentation to see if accuracy on synthesized images improves, particularly for closed-loop-like trajectories as in~\ref{fig:traj_mesh}. 

\section{Acknowledgments}

The results presented in this paper have been achieved during a research stage at the European Space Agency (ESA). The views expressed in this paper can in no way be taken to reflect the official opinion of the European Space Agency.

The authors thank the TEC-ECG section of ESA for all the inspiring conversations and teachings that helped bring this work to life. In particular, we want to thank Paul Duteïs for his generous help and continuous availability, as well as Massimo Casasco for his trust and the opportunity to pursue this work. 

%
%
\begingroup
  \small              
  \bibliographystyle{plain}
  \bibliography{cleaned_library_241216}
\endgroup

\end{document}